%% file: main.tex
\documentclass[10pt,twocolumn,letterpaper]{article}

\usepackage{3dv}
\usepackage{times}
\usepackage{epsfig}
\usepackage{graphicx}
\usepackage{amsmath}
\usepackage{amssymb}

\usepackage{gensymb}

\usepackage{subcaption}
\usepackage{enumitem}

\usepackage[pagebackref=true,breaklinks=true,colorlinks,bookmarks=false]{hyperref}

\threedvfinalcopy %

\def\threedvPaperID{67} %

\ifthreedvfinal\fi

\input{tex/00_abbrev_and_macros}

\makeatletter
\def\blfootnote{\gdef\@thefnmark{}\@footnotetext}
\makeatother

\begin{document}

\title{Leveraging MoCap Data for Human Mesh Recovery}

\author{
Fabien Baradel$^{*}$
\and
Thibault Groueix$^{*}$
\and
Philippe Weinzaepfel
\and
Romain Br\'egier
\and
Yannis Kalantidis \hspace{10mm}  Gr\'egory Rogez \vspace{1mm}\\
{NAVER LABS Europe}\\
{\tt\small firstname.lastname@naverlabs.com}
} 

\maketitle
\thispagestyle{empty}

\blfootnote{$^{*}$ indicates equal contribution. Thibault Groueix is now at Adobe.}
\input{tex/00_abstract}

\input{tex/01_introduction}
\input{fig/synthetic_renderings}
\input{tex/02_related}

\input{tex/03_method}

\input{tex/04_experiments}

\input{tex/05_conclusions}

{\small
\bibliographystyle{ieee_fullname}
\bibliography{egbib}
}

\input{tex/06_supp_mat}

\end{document}

%% file: tex/00_abbrev_and_macros.tex
\usepackage[dvipsnames, table]{xcolor}
\usepackage{xspace}
\usepackage{xcolor}
\usepackage{diagbox}
\usepackage{bm}
\usepackage{amssymb}
\usepackage{booktabs}

\usepackage{multirow}
\usepackage{caption}

\definecolor{Gray}{gray}{0.9}
\definecolor{lightgray}{gray}{0.94}

\newcommand{\smpltransformer}{PoseBERT\xspace}
\newcommand{\poseformer}{\smpltransformer}
\newcommand{\posebert}{\smpltransformer}
\newcommand{\sspin}{MoCap-SPIN\xspace}

\newcommand{\errorgain}[1]{\color{OliveGreen}{($\downarrow$ #1)}}

\makeatletter
\DeclareRobustCommand\onedot{\futurelet\@let@token\@onedot}
\def\@onedot{\ifx\@let@token.\else.\null\fi\xspace}

\def\eg{\emph{e.g}\onedot} 
\def\ie{\emph{i.e}\onedot}

\def\etal{\emph{et al}\onedot}
\makeatother

\newcommand{\vx}{\mathbf{x}}
\newcommand{\vy}{\mathbf{y}}

\newcommand{\vxsynth}{\mathbf{s}}

\renewcommand{\paragraph}[1]{\vspace{0.02cm}\noindent\textbf{#1}}

%% file: tex/00_abstract.tex
\begin{abstract}
\noindent
Training state-of-the-art models for human body pose and shape recovery from images or videos requires datasets with corresponding annotations that are really hard and expensive to obtain.
Our goal in this paper is to study whether poses from 3D Motion Capture (MoCap) data can  be used to improve image-based and video-based human mesh recovery methods.
We find that fine-tune image-based models with synthetic renderings from MoCap data can increase their performance, by providing them with a wider variety of poses, textures and backgrounds. In fact, we show that simply fine-tuning the batch normalization layers of the model is enough to achieve large gains. 
We further study the use of MoCap data for video, and introduce \smpltransformer, a transformer module that directly regresses the pose parameters and is trained via masked modeling.
It is simple, generic and can be plugged on top of any state-of-the-art image-based model in order to transform it in a video-based model leveraging  temporal information.
Our experimental results show that the proposed approaches reach state-of-the-art performance on various datasets including 3DPW, MPI-INF-3DHP, MuPoTS-3D, MCB and AIST. Test code and models will be available soon.
\end{abstract}

%% file: tex/01_introduction.tex
\begin{figure}
\centering
\includegraphics[width=\linewidth]{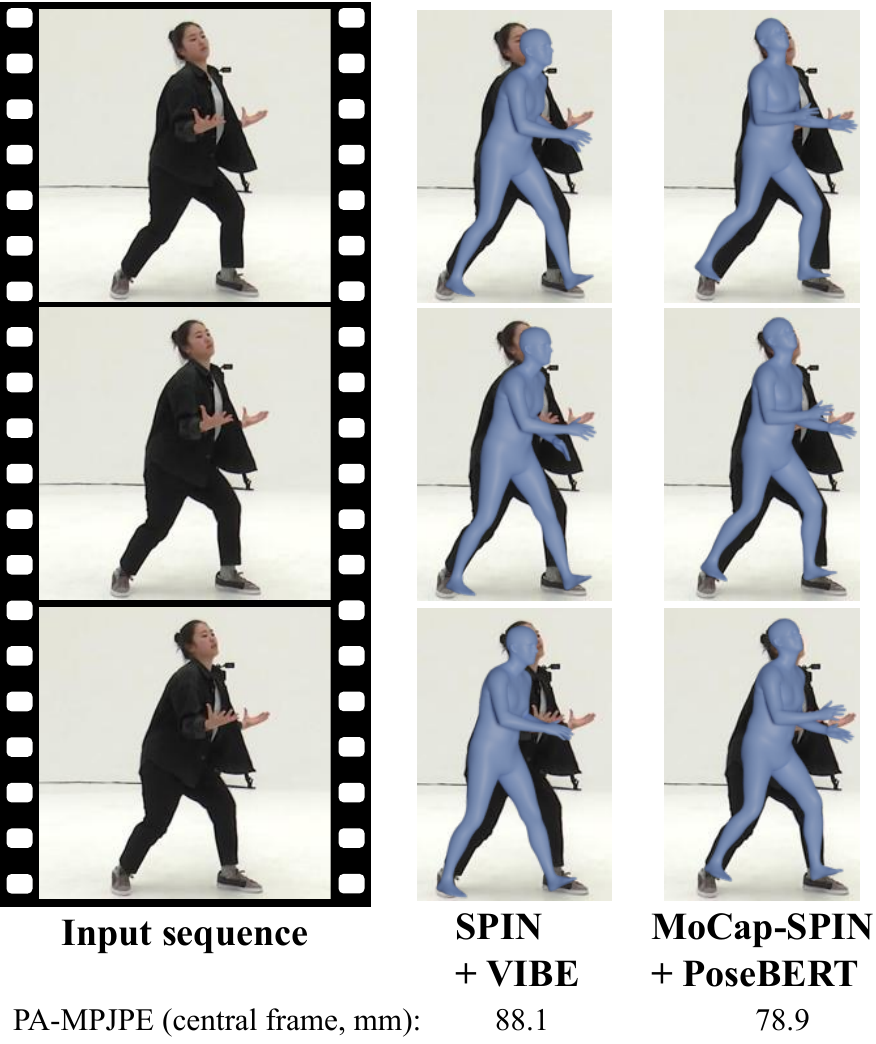} \\[-0.25cm]
\caption{\label{fig:teaser}
\textbf{
Qualitative comparison between SPIN+VIBE baseline~\cite{vibe} and the proposed method.} We achieve state-of-the-art results for both image- and video-based pose estimation on a large variety of datasets (here a video sequence from AIST~\cite{aist}) by leveraging MoCap data.
}

\vspace{-0.5cm}

\end{figure}

\section{Introduction}
\label{sec:introduction}

State-of-the-art methods that estimate 3D body pose and shape given an image~\cite{hmr,nbf,spin,pose2mesh} or a video~\cite{vibe,meva} have recently shown impressive results. A major challenge when training models for in-the-wild human pose estimation is data: collecting large sets of training images with ground-truth 3D annotations is cumbersome as it requires setting up IMUs~\cite{3dpw}, calibrating a multi-camera system~\cite{huang2017} or considering static poses~\cite{mc3dv}.  
In practice, only 2D information such as 2D keypoint locations or semantic part segmentation can be manually annotated.
Current methods therefore leverage this in-the-wild data with partial ground-truth annotations by defining their losses on the 2D reprojection of the 3D estimation~\cite{hmr}, by running an optimization-based method~\cite{smplify} beforehand and curate the obtained ground-truth~\cite{unite} or by running the optimization inside the training loop~\cite{spin}. 
This lack of annotated real-world data is even more critical for videos, making difficult the use of recent temporal models such as transformers~\cite{transformer} which are known to require large datasets for training.

At the same time, Motion Capture (MoCap), widely employed in the video-game and film industry, offers a solution to create large corpus of motion sequences with accurate ground-truth 3D poses. 
Recently, several of these MoCap datasets have been unified into the large AMASS dataset~\cite{amass}. Importantly, all the considered MoCap sequences were converted into realistic 3D human meshes represented  by a rigged body model, concretely SMPL~\cite{smpl},  a  differentiable  parametric  model employed by most state-of-the-art human mesh recovery methods~\cite{smplify,hmr,nbf,spin,pose2mesh}. 
In previous works, MoCap data have been used to train a discriminator for image-based~\cite{hmr} or video-based~\cite{vibe} models. In this case, these data are only used to force the model to predict a realistic output, without improving its capability to better estimate the poses by seeing more diverse examples. Another way to exploit MoCap data consists in rendering it, applying a texture and adding a background image as in~\cite{surreal}, to generate synthetic images with ground-truth annotations. However, the inherent domain shift between synthetic and real-world images has limited  the use of such data to pretraining~\cite{bodynet}.
\newline
\indent In this paper, we study how MoCap data can be used in order to improve methods that estimate SMPL parameters in images or videos.
For image-based models, we show that fine-tuning SPIN~\cite{spin}, a state-of-the-art method, while adding renderings from AMASS to its original training data of real images, improves its performance. 
We hypothesize this is a consequence of the model being exposed to a wider and more diverse set of poses. For added realism, we follow SURREAL~\cite{surreal} and render textured SMPL models on top of random background images, increasing the invariance of the model to texture and background change. 
Surprisingly, we show that fine-tuning only the batch normalization layers of the pretrained SPIN model suffices to get state-of-the-art performance for image-based models. 
\newline
\indent We further utilize MoCap data for learning better video models, and introduce \posebert, a transformer-based model tailored for pose estimation that directly regresses the pose parameters. \posebert takes as input estimated SMPL pose parameters and can be learned on the AMASS dataset using masked modeling, similar to BERT~\cite{bert}. This module can be plugged on top of any state-of-the-art image-based model in order to transform it into a video-based model. A qualitative comparison between SPIN and the proposed model is shown in~\ref{fig:teaser}.
In summary, our contributions are:
\begin{itemize}[noitemsep,topsep=-2pt]
  \item We leverage MoCap data to make image-based models more robust by fine-tuning using synthetic renderings.
  \item We introduce \posebert, a transformer model that regresses pose parameters and is trained using MoCap data via masked modeling. It is a plug-and-play unit that allows any image-based model to leverage temporal context.
  \item We exhaustively ablate all aspects of our methods and present consistent gains over the state of the art on several challenging  datasets captured in the wild.
\end{itemize}

%% file: fig/synthetic_renderings.tex
\begin{figure*}[th!]%
\centering
\newlength{\synthfigwidth}
\setlength{\synthfigwidth}{0.12\linewidth}
\begin{subfigure}{\textwidth}
\hspace{11pt}
\includegraphics[height=\synthfigwidth]{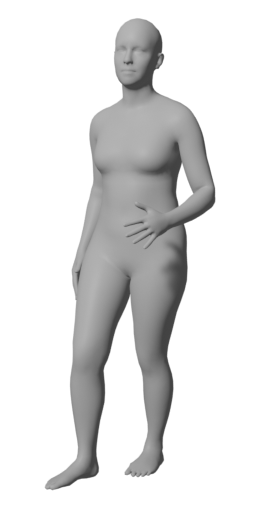}
\hspace{11pt}
\includegraphics[width=\synthfigwidth]{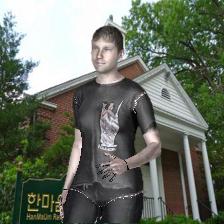}
\includegraphics[width=\synthfigwidth]{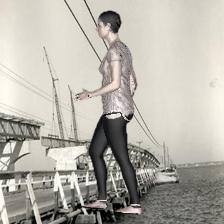}
\includegraphics[width=\synthfigwidth]{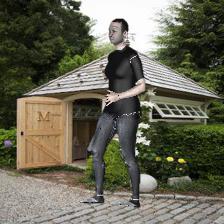}
\includegraphics[width=\synthfigwidth]{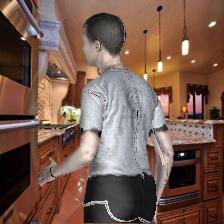}
\includegraphics[width=\synthfigwidth]{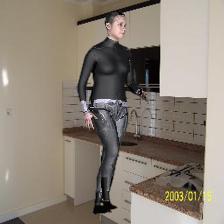}
\includegraphics[width=\synthfigwidth]{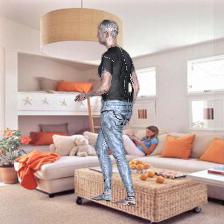}
\includegraphics[width=\synthfigwidth]{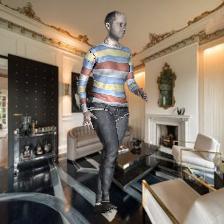}
\vspace{-0.15cm}
\caption{\label{fig:synth_amass} Different possible renderings for a ground-truth pose from AMASS~\cite{amass}. The same pose can be rendered with random camera viewpoint, texture, background to generate diverse synthetic training data.}
\end{subfigure}
\begin{subfigure}{\textwidth}
\setlength{\synthfigwidth}{0.115\linewidth}
\includegraphics[width=\synthfigwidth]{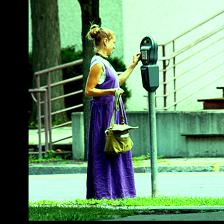}
\includegraphics[width=\synthfigwidth]{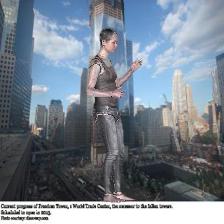} 
\hspace{3pt}
\includegraphics[width=\synthfigwidth]{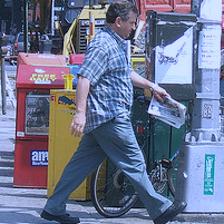}
\includegraphics[width=\synthfigwidth]{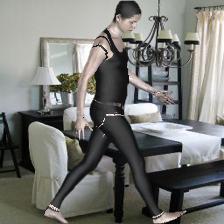}
\hspace{3pt}
\includegraphics[width=\synthfigwidth]{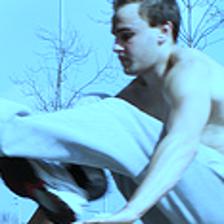}
\includegraphics[width=\synthfigwidth]{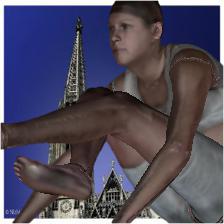}
\hspace{3pt}
\includegraphics[width=\synthfigwidth]{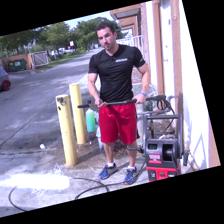}
\includegraphics[width=\synthfigwidth]{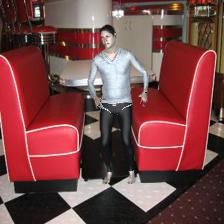} \\[-0.4cm]
\vspace{-0.15cm}
\caption{\label{fig:synth_pairs} Pairs of augmented real images from COCO~\cite{coco}, LSPE~\cite{lsp}, MPII~\cite{mpii} and corresponding synthetic renderings depicting the same pose with a different background and body appearance.}
\end{subfigure}
\vspace{-0.3cm}
\caption{\label{fig:synth_examples} \textbf{Examples of synthetic human renderings.}}
\vspace{-0.5cm}
\end{figure*}

%% file: tex/02_related.tex
\section{Related work}
\label{sec:related}

We review related work on the estimation of the SMPL parameters from images or videos as well as the use of synthetic data for human pose and shape estimation.

\paragraph{SMPL from images.}
SMPLify~\cite{smplify} first introduced an optimization-based method to find the SMPL parameters that best explain a set of detected 2D keypoints by leveraging various priors, but it remains sensible to the initialization.
Since then, most deep learning methods~\cite{nbf,hmr,spin,graphcmr,pose2mesh} process image crops around a person of interest to directly estimate these SMPL parameters. 
In order to handle real-world images, they are trained on a mix of real images where annotations are limited to 2D keypoints, and synthetic or MoCap images for which 3D poses are available.
Losses, typically applied on the difference between SMPL 2D or 3D keypoint predictions and the corresponding annotations, can also be applied on the vertices~\cite{spin,graphcmr} or on texture correspondences~\cite{dct,texturepose}.
In some works~\cite{hmr,texturepose}, an additional discriminator ensures the realism of the predicted SMPL parameters.
SPIN~\cite{spin} combines deep learning-based method with optimization-based approaches by using the optimization stage to refine the prediction made by the network, which is later used in upcoming epochs.
In I2L-MeshNet~\cite{i2lmeshnet}, vertices' locations are estimated with heatmaps for each mesh vertex coordinate instead of directly regressing the parameters while in Pose2Mesh~\cite{pose2mesh}, a 2D pose is first lifted to 3D to obtain a coarse mesh which is then iteratively refined. Both methods are trained on images where SMPL pseudo-ground-truth was obtained from 2D keypoints using SMPLify~\cite{smplify}, which can result in inaccurate 3D data. Instead, we employ the final estimations in the SPIN training procedure to train with direct supervision and fully exploit synthetic data.

\paragraph{SMPL from videos.}
While Arnab \etal~\cite{videokinetics} proposed an optimization-based strategy to handle human pose estimation in videos, recent methods are mostly based on deep learning~\cite{vibe,thmmr,Jiang_2021_CVPR}.
In HMMR~\cite{hmmr}, features from consecutive frames are fed to a 1D temporal convolution, while VIBE~\cite{vibe} uses recurrent neural network, namely Gated Recurrent Unit (GRU), together with a discriminator at the sequence level.
The network is trained on different in-the-wild videos and losses are similar to the ones employed for images and described above, \ie, mainly applied on keypoints.
A similar architecture with GRU is used in TCMR~\cite{tcmr}, except that 3 independent GRUs are used and concatenated, one in each direction and one bi-directional in order to better leverage temporal information.
MEVA~\cite{meva} estimates motion from videos by  also extracting temporal features using GRUs and then estimates the overall coarse motion inside the video with Variational Motion Estimator (VME).
Recently, Pavlakos \etal~\cite{thmmr} have proposed to use a transformer architecture~\cite{transformer} in a concurrent work.
To obtain training data, \ie, in-the-wild videos annotated with 3D mesh information, they use the smoothness of the SMPL parameters over consecutive frames to obtain pseudo-ground-truth. In terms of architecture, the transformer is used to leverage temporal information by modifying the features.
We also consider a transformer architecture but, in our case, it is directly applied to the sequences of SMPL parameters. It has the great advantage of being directly trainable on MoCap data and pluggable to any image-based method.

\paragraph{Learning with synthetic data.}
Employing synthetic training images is a standard strategy to overcome the lack of large scale annotated data in computer vision~\cite{GaidonLP18}. This is particularly the case for human 3D pose estimation as it is not possible to accurately annotate 3D information on a large corpus of in-the-wild images~\cite{ChenWLSWTLCC16,RogezS16,surreal}. Usually, a sim2real domain  gap arises  between  synthetic  and real examples that needs to be carefully handled to ensure generalization.  For example, Chen \etal~\cite{ChenWLSWTLCC16} employed computer generated images of people to train a 3D pose estimation network with a domain mixer adversarially trained to discriminate between real and synthetic images, and therefore handle domain adaptation. 
More recently, SimPose~\cite{simpose} proposed to train a model for 2.5D pose estimation using a mixed of synthetic and real data where 2D losses are applied to real data while 2.5D and 3D losses are applied to synthetic data only, which may affect generalization.
Kundu \etal~\cite{kundu2020unsupervised} proposed to estimate an auto-encoder of 3D pose, together with a network that processes real images and output 2D poses. Losses are also applied on the interleaved network to allow training a model to estimate 3D pose in the wild. However, this does not allow the image feature extractor to see more variability in terms of poses. Closer to us is the work by Varol \etal~\cite{surreal} who apply real textures on renderings of the SMPL parametric model completed with real background images to generate a large and varied dataset. In a subsequent work, they trained a network for volumetric body estimation~\cite{bodynet} by pretraining on this large synthetic data and fine-tuning on more limited real data.
Doesh \etal~\cite{sim2real} show that motion through the optical flow in synthetic videos from~\cite{surreal} allows to bridge the sim2real gap for 3D human pose estimation.
In this paper, we also leverage SMPL renderings but fully exploit the synthetic data, \ie, without limiting its use to pretraining.

%% file: tex/03_method.tex
\section{MoCap data for human mesh recovery}
\label{sec:method}

In this section we present two ways of using MoCap data for 3D mesh estimation in images and videos.
We first explore MoCap data for regularizing image-based models via fine-tuning using synthetic renderings %
(Section~\ref{sec:method_fine-tuning}). 
We then introduce \posebert, a pose estimation-oriented transformer architecture that is trained on MoCap data and enables the model to leverage temporal context (Section~\ref{sec:method_temporal}).

\subsection{Improving image-based pose estimation}
\label{sec:method_fine-tuning}

Existing datasets with ground-truth pose annotations for real-world images are not large enough to capture the variability of images/video sequences that can be encountered at test time; this can limit the generalization capabilities of a pose estimation method. We argue that one way to mitigate this issue is through the use of renderings from MoCap data. We leverage such synthetic data to regularize a strong image-based model like SPIN~\cite{spin}. SPIN is a 3D pose estimation model pretrained on datasets with paired RGB and 2D and/or 3D pose annotations. We propose to fine-tune this model using synthetic renderings and therefore expose it to a more diverse set of poses, viewpoints, textures and background changes from the ones seen during pretraining.

Following~\cite{hmr,spin}, we use a deep convolutional neural network encoder $f$ for feature extraction from image crops followed by an iterative regressor $r$. Let $\vx \in \mathrm{R}^d$ be an input crop, $\phi_\vx = f(\vx)$ be the extracted feature and $r(\phi_\vx) = \{ \Theta_\vx, \Pi_\vx \}$
the regressed prediction for the $d_\theta$-dimensional body model parameters $\Theta_\vx$ and the $d_\pi$-dimensional camera parameters $\Pi_\vx$. The regressor $r$ is  initialized with the mean pose parameters $\Theta_\text{mean}$ and is run for a number of $N$ iterations. Similar to~\cite{spin}, our network uses the 6D representation proposed in~\cite{rotation} for 3D joints orientation.

\begin{figure}[t]
\centering
\includegraphics[width=.95\linewidth]{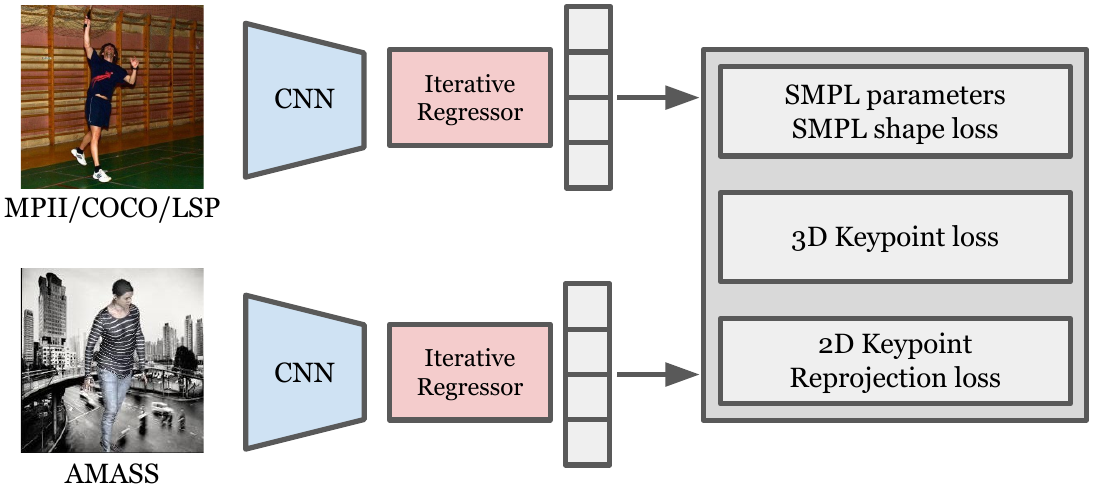} \\[-0.2cm]
\caption{\label{fig:finetuning} \textbf{Regularizing image-based human mesh recovery model.} We leverage synthetic renderings of MoCap data from AMASS~\cite{amass} together with real images to regularize  SPIN~\cite{spin}. To fine-tune SPIN, we train the batch-normalization layers~\cite{batchnorm} of the CNN backbone.}
\vspace{-0.6cm}
\end{figure}

An overview of the fine-tuning process is depicted in Figure~\ref{fig:finetuning}. We sample poses from  AMASS~\cite{amass}, a large collection of MoCap data, and use them together with real-world images for fine-tuning the SPIN model. Given that we are starting from a state-of-the-art-model, apart from fine-tuning all of the backbone parameters, we also experiment with only fine-tuning a subset. Specifically, and motivated by a recent work~\cite{frankle2020training}, we experiment with only fine-tuning the parameters of the batch normalization~\cite{batchnorm} layers, \ie the affine transformation in each such layer, together with the running statistics.

Providing both real and synthetic data is crucial since fine-tuning only using synthetic data would hurt performance due to the synthetic/real data domain shift.
We therefore fine-tune the SPIN model with batches that partially contain the same real data as used in SPIN during training, together with our synthetic renderings from AMASS poses. We initialize the weights with the released model from SPIN, and use direct supervision: for synthetic data we use the corresponding ground-truth SMPL parameters, for real data we use the final fits from the SPIN training procedure as pseudo ground-truth.
We refer to a SPIN model fine-tuned this way as \textit{\sspin}.

\paragraph{Rendering synthetic humans.}
To generate a synthetic rendering of a human in a given pose, we render the corresponding  SMPL~\cite{smpl} model with a random texture from the SURREAL dataset~\cite{surreal}, using a random background image from LSUN training set~\cite{lsun} to provide further data augmentation.
Figure~\ref{fig:synth_examples} shows examples of such renderings.

We randomly sample SMPL pose and shape parameters from AMASS~\cite{amass}, which provides us with a great source of diverse poses. 
Camera 3D orientation is sampled uniformly considering a Tait-Bryan parametrization ($\pm180\degree{}$ yaw, $\pm 45\degree$ pitch, $\pm15\degree{}$ roll, with yaw and roll axes horizontal when considering the identity rotation) to model typical variability observed in real data.
Synthetic renderings are cropped around the person in a similar manner as for real images, based on the location of 2D joints of the model. We model the fact that people are not always entirely visible in real crops by considering only upper body keypoints up to the hips or the knees in $20\%$ of the cases. Figure~\ref{fig:synth_amass} illustrates the variability of renderings given a single pose.

In order to verify that sampling diverse poses from large MoCap datasets is preferable over creating renderings of the poses that 
already appear in the commonly used datasets like COCO~\cite{coco}, MPII~\cite{mpii}, LSP and LSPE~\cite{lsp}, we further experiment with rendering poses from such datasets as well. Although restricted to the set of training poses, augmenting the dataset with synthetic renderings of the same poses can offer greater variability in textures and backgrounds. We use the pseudo-ground-truth (\ie, final fits from SPIN) pose, shape and camera parameters from the real data to render synthetic humans following the process presented above. Examples of pairs of real data and synthetic renderings are illustrated in Figure~\ref{fig:synth_pairs}.

\subsection{Capturing temporal context with transformers}
\label{sec:method_temporal}

\begin{figure}[t]
\centering
\includegraphics[width=\columnwidth]{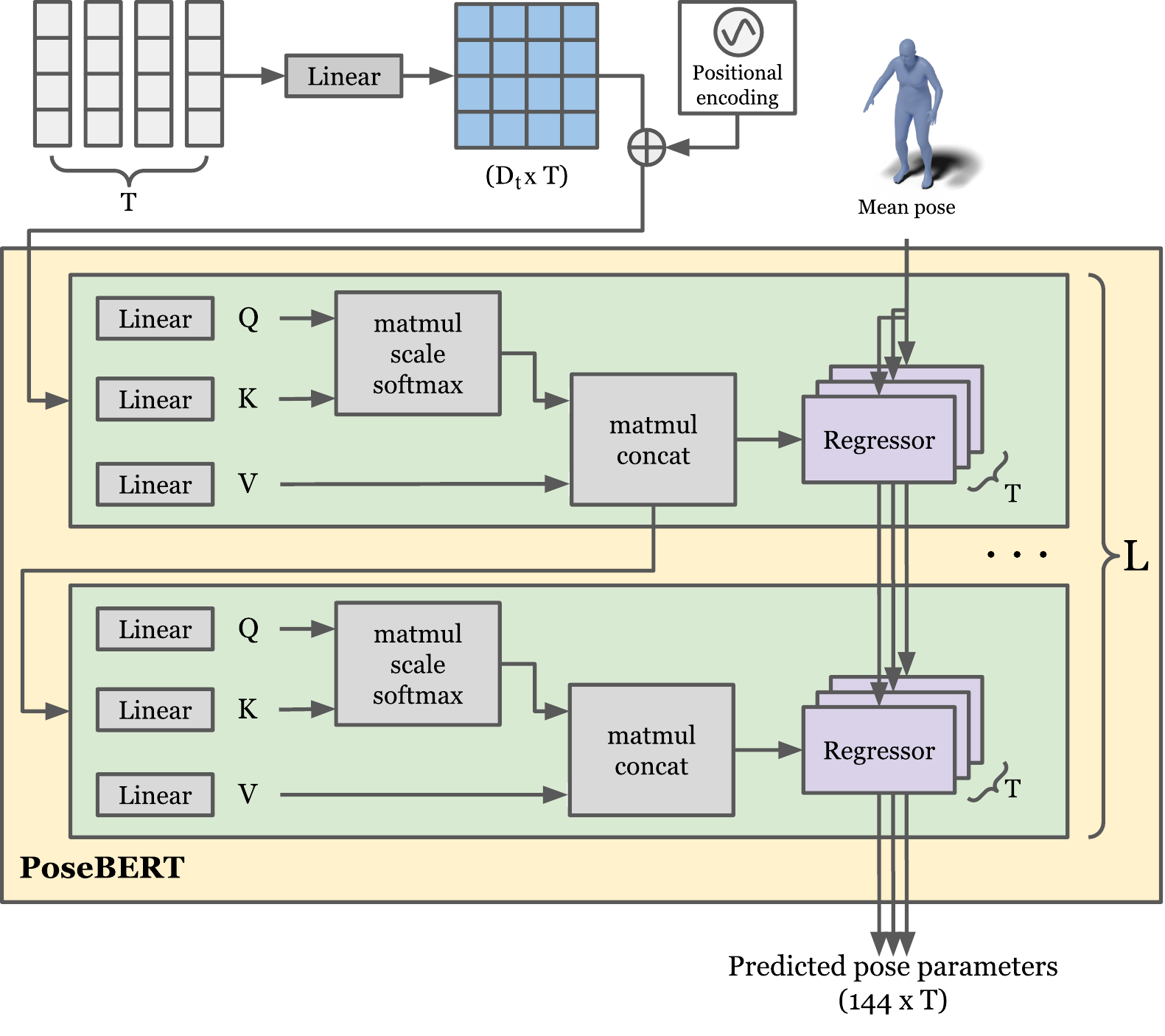} \\[-0.3cm]
\caption{\label{fig:transformer}\textbf{The \posebert architecture}. Both input and output are SMPL pose parameters for a temporal sequence of $T$ poses. \smpltransformer basic block is repeated $L$ times. The regressor parameters are shared across the $T$ inputs and the $L$ blocks. We regress the pose starting from the mean pose.
}
\vspace{-0.5cm}
\end{figure}

In this section, we present \textbf{\smpltransformer}, a transformer-based architecture that regresses \emph{directly} the SMPL parameters for every pose in a temporal sequence. Although concurrent works have also utilized temporal models based on GRUs~\cite{vibe,meva,tcmr} or transformers~\cite{thmmr} for human mesh recovery, they utilize context to enhance the visual features, and still require an iterative regressor on top of the temporal model for regressing the SMPL parameters. 

\paragraph{The basic \smpltransformer block.}
Figure~\ref{fig:transformer} illustrates the architecture of \smpltransformer. The input is $T$ vectors, corresponding to a temporal sequence of poses of length $T$ and the output is a sequence of SMPL pose parameters predictions, one for each frame. 
We propose to incorporate the regressor as part of the basic transformer block: we replace the feed-forward network that follows the multi-headed dot-product attention with a regressor $r$, similar to the one used in~\cite{hmr}. This gives us a main advantage: at every layer of \smpltransformer, we dynamically attend to parts of the input sequence and refine the regressed SMPL pose prediction accordingly.

Since we are learning from synthetic data, pose ground-truth is available. 
Despite the fact that the transformer only sees ground-truth poses during training, as we experimentally show in Section~\ref{sec:experiments}, the learned model can generalize to noisy poses during testing, \ie, when the input is instead the pose predictions from any image-based model.

The basic block of \smpltransformer is repeated $L$ times. 
Inside each block, the input is first fed to a multi-head scaled dot-product attention mechanism and then to a regressor MLP. For the latter we use the architecture from~\cite{hmr} and share the regressor parameters across the $T$ inputs.  Although the regressor can be iterative as in~\cite{hmr}, given that this is a process that happens at every layer, \ie, $L$ times, we find that one iteration is enough (see ablation in Section~\ref{sec:experiments}). Moreover, we experiment with versions of \smpltransformer where the regressor parameters are shared across the $L$ blocks, thus reducing the number of learnable parameters.

Like the original transformer~\cite{transformer}, we use layer normalization before self-attention modules. For the first layer only, and similarly to~\cite{transformer}, we add a 1-D positional encoding to the input which is learned from scratch. We also initialize the first layer regressor with the mean pose $\Theta_\text{mean}$. Finally, we first learn a linear projection to $D_t$ dimensions before we feed the input to the transformer; the choice of this parameter directly influences the size of the model.

\begin{figure}[t]
\centering
\includegraphics[width=\columnwidth]{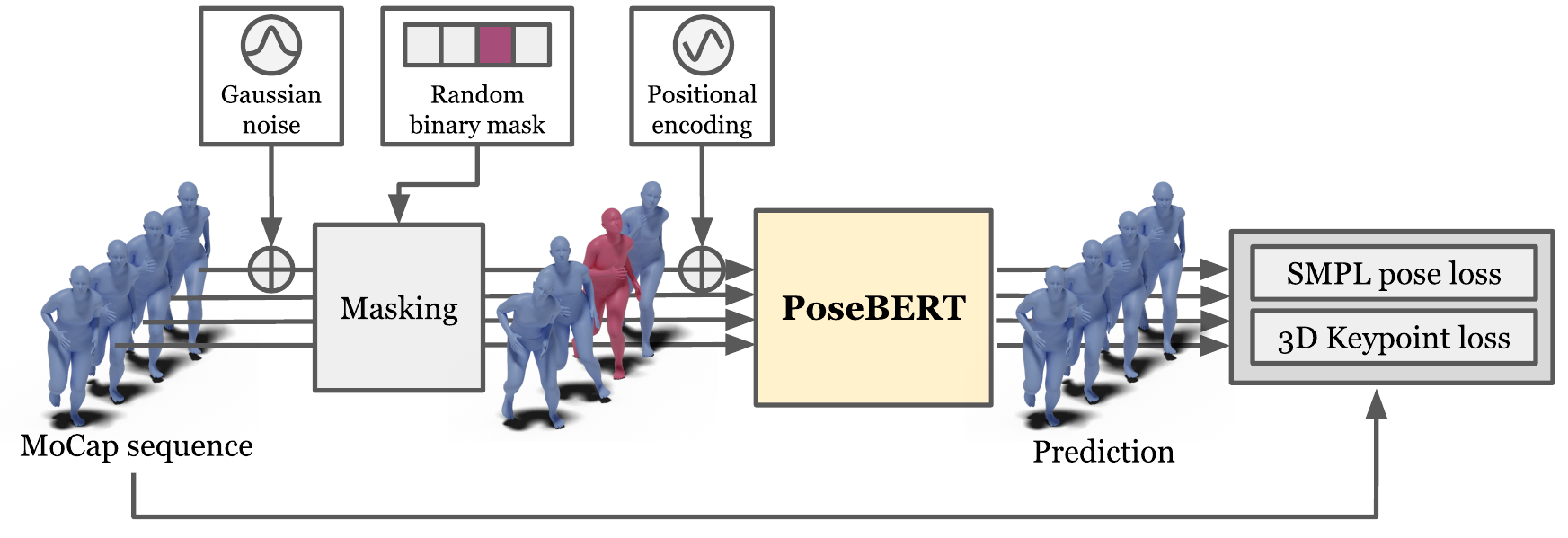} \\[-0.2cm]
\vspace{-0.15cm}
\caption{\label{fig:temporal}\textbf{Learning \smpltransformer with masked modeling.} To allow the model to learn temporal dynamics and similar to~\cite{bert} we \emph{mask} part of the input and either replace it with a learnable masking token or a random pose. The mask is a random $T$-dimensional binary vector that specifies which timestamps will be masked.
}
\vspace{-0.6cm}
\end{figure}

\paragraph{Learning \smpltransformer parameters.}
Looking at context can help to correct errors of models based on single images. By inputting a sequence of poses to \poseformer,  we want to be able to learn \emph{temporal} dynamics. To do so, we utilize the masked modeling task, one of the self-supervised tasks that the now ubiquitous BERT~\cite{bert} model uses to learn language models. We also add random Gaussian noise to the input.
Figure~\ref{fig:temporal} shows an overview of the training process. The inputs are first masked and noise is added. This is then given to  \poseformer as an input. \poseformer outputs are then compared to the original clean inputs and similarly to the image-based model; for every timestep there is a loss on the SMPL pose parameters as well as a 3D keypoint loss.

In masked modeling, part of the input is masked before it is fed to the model. The correct output is expected to be recovered using the rest of the sequence.
In practice, and for each input sequence, we create a $T$-dimensional random binary vector, where each timestamp has a probability $m/T\%$ to be masked, where $m$ is an hyperparameter that controls the number of frames of the input that will be masked.
For the $m$ timesteps that are randomly chosen to be masked, we replace the input pose with a learnable masked token and prohibit the self-attention module to attend to those timesteps.
We ablate $m$ in the experiments.

%% file: tex/04_experiments.tex
\section{Experiments}
\label{sec:experiments}

After presenting datasets and metrics in Section~\ref{sub:data}, we perform extensive evaluations and ablations of \sspin (Section~\ref{sub:xpsspin}) and of \posebert (Section~\ref{sub:xpposebert}). We finally compare to the state of the art in Section~\ref{sub:xpsota}

\begin{table*}[t]
\centering
\resizebox{\linewidth}{!}{
\begin{tabular}{lccc|cc|cc|cc|c}
\toprule
      & \multicolumn{3}{c|}{3DPW~\cite{3dpw}} & \multicolumn{2}{c|}{ MuPoTS-3D~\cite{mupots} } & \multicolumn{2}{c|}{ AIST~\cite{aist} } & \multicolumn{2}{c|}{ MPI-INF-3D~\cite{mpiinf} } & MCB~\cite{mc3dv} \\
      & MPJPE $\downarrow$ & $\mathcal{E}$ $\downarrow$ & MPVPE $\downarrow$ & MPJPE $\downarrow$ & $\mathcal{E}$ $\downarrow$ & MPJPE $\downarrow$ & $\mathcal{E}$ $\downarrow$ & MPJPE $\downarrow$ & $\mathcal{E}$ $\downarrow$ & MPJPE $\downarrow$ \\
    \midrule %
     \textbf{SPIN~\cite{spin}} & 97.2 & 59.6 & 116.8 & 154.6 & 83.0 & 126.2 & 76.2 & 104.3 & 68.0  & 155.4 \\
     \textit{fine-tune all parameters} & 95.1 & 58.6 & 112.1 & 154.8 & 82.8 &129.6 & 76.6  & 102.7 & \textbf{66.6} & 150.6 \\
    \textit{fine-tune batch-norm layers} & 94.8 & 58.1 & 111.6 & 153.7 & 82.5 & 126.4 & 76.0 & 102.4 & 67.1 & 149.7 \\
    \midrule
     \textbf{Using synthetic data } & & & & & \\
    \textit{\ \ \ + renderings of SPIN data} & 93.5 & 58.6 & 109.7 &\textbf{152.2} & 82.1 & \textbf{123.8} & 76.0 & \textbf{98.0} & 67.4 & 150.0 \\
    \textit{\ \ \ + renderings of MoCap data} (\sspin) & \textbf{90.8} & \textbf{55.6} & \textbf{105.0} & 152.3 & \textbf{81.0} & 125.1 &  \textbf{75.7} & 100.8 & 66.7 & \textbf{145.0} \\
\bottomrule
\end{tabular}
}

\vspace{-0.3cm}

\caption{
{
\textbf{Fine-tuning the pretrained SPIN model.}  For brevity, we denote the PA-MPJPE metric as  $\mathcal{E}$. Top section presents sanity checks, \ie, when fine-tuning the model with the same data used during pretraining (denoted as SPIN data). Bottom section presents results further adding synthetic renderings during fine-tuning, either coming from the SPIN data, or from AMASS (\sspin). Results in the bottom part are obtained when fine-tuning only the batch normalization layers.}
}
\label{tab:single_image_ablation}
\vspace{-0.5cm}
\end{table*}

\subsection{Datasets, training and metrics}
\label{sub:data}

\noindent \textbf{MoCap data.} We use AMASS~\cite{amass}, which is a collection of numerous Motion Capture datasets in a unified SMPL format, representing more than 45 hours of recording.

\noindent \textbf{Real-world data (SPIN data).} When fine-tuning the image-based model, we use the same real-world training data as SPIN~\cite{spin}, namely 2D pose estimation datasets like COCO~\cite{coco}, LSP and LSPE~\cite{lsp} and MPII~\cite{mpii}, as well as a 3D pose estimation dataset, MPI-INF-3DHP~\cite{mpiinf}. We were not able to use Human 3.6M~\cite{h36} due to license restriction.

\noindent \textbf{Training.}
We first train \sspin by fine-tuning the BN layers of a pre-trained SPIN model on SPIN data plus AMASS. We then train \posebert solely on MoCap data.

\noindent \textbf{Test datasets and metrics.}
For evaluation, we use the 3DPW~\cite{3dpw} test set, the MPI-INF-3DHP~\cite{mpiinf} test set, the MuPoTS-3D dataset~\cite{mupots} and the AIST dataset~\cite{aist} that contains more challenging poses from people dancing. For image-based models, we also use the Mannequin Challenge Benchmark (MCB)~\cite{mc3dv}. We report the mean per-joint error (MPJPE) before and after procrustes alignment (PA-MPJPE) in millimeters (mm). For 3DPW, following the related work, we also report the mean per-vertex position error (MPVPE). To measure the jittering of the estimations on the video datasets (3DPW, MPI-INF-3DHP, MuPoTS-3D, AIST), we follow~\cite{hmmr} and report the acceleration error, measured as the average difference between ground-truth and predicted acceleration of the joints.

\subsection{Evaluation of \sspin}
\label{sub:xpsspin}

We first evaluate and ablate \sspin, our approach to fine-tune a state-of-the-art image-based model using synthetic renderings of MoCap data. 
In Table~\ref{tab:single_image_ablation} we report results on five datasets after fine-tuning the publicly available pretrained model of SPIN\footnote{\url{https://github.com/nkolot/SPIN}}. 
In the top section we report sanity checks, \ie, performance when fine-tuning the model with the same data used during pretraining (denoted as SPIN data).
During early experiments, fine-tuning all model parameters using both real and synthetic renderings made the model quickly overfit and performance on the test datasets started to diverge as training progressed. To mitigate overfitting, we explored fine-tuning only subsets of the backbone parameters, \eg only the regressor, only the last layers or to freeze the batch normalization layers and statistics. 
Inspired by a recent paper~\cite{frankle2020training} we also experimented with training only the batch normalization layer parameters, freezing the rest of the network. 

Surprisingly, even without additional synthetic data, fine-tuning only the batch normalization layers provides a consistent boost on most datasets and metrics (second row of Table~\ref{tab:single_image_ablation}).
In fact, fine-tuning only the batch normalization layer parameters was the setting that enabled us to fine-tune using synthetic data. As shown in the bottom section of Table~\ref{tab:single_image_ablation}, we observe large gains in performance across all datasets in both cases.
This behavior about batch-norm statistics has already been shown by ~\cite{schneider2020improving}.
When the renderings come from poses sampled from SPIN data, gains show that learning with diverse backgrounds and textures make the model more robust.  The gains are however even larger when the renderings come from poses sampled from AMASS; sampling a more diverse set of poses seems to directly affect generalization performance. For the rest of the experiments, we will refer to \sspin as the SPIN model whose batch normalization parameters were fine-tuned using both real data and MoCap renderings.

\begin{figure}
\centering
\includegraphics[width=\columnwidth]{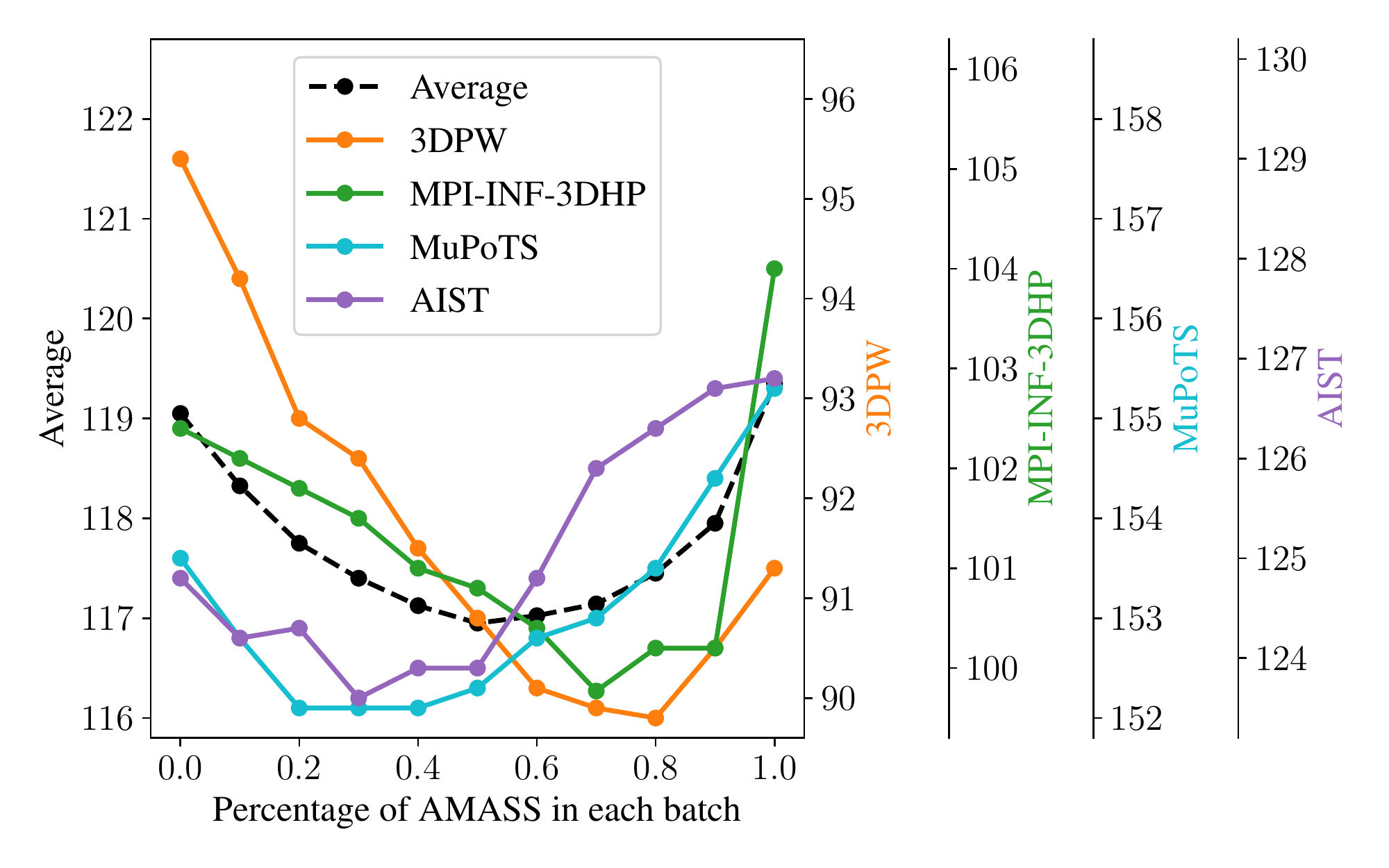} \\[-0.4cm]
\vspace{-0.15cm}
\caption{\textbf{Varying the percentage of synthetic data}. We report the MPJPE metric on four datasets when varying the percentage of AMASS rendering at each batch. We further plot the  average across the four datasets (dashed black curve); the best performance overall is achieved when the batch is equally split between real and synthetic data.}
\label{fig:mocap_percentage}
\vspace{-0.6cm}
\end{figure}
 
\paragraph{Ablating the balance between real and synthetic data in each batch.} 
We study the impact of varying the percentage of synthetic renderings inside each batch. Results are shown in Figure~\ref{fig:mocap_percentage}.
On all datasets, we observe a U shape curve as the percentage of synthetic data increases, meaning that the optimal value is not an extreme value (neither only real data nor only synthetic). Using half of synthetic data is a good compromise on all datasets.
 
\begin{figure}
\centering
\includegraphics[width=0.5\columnwidth]{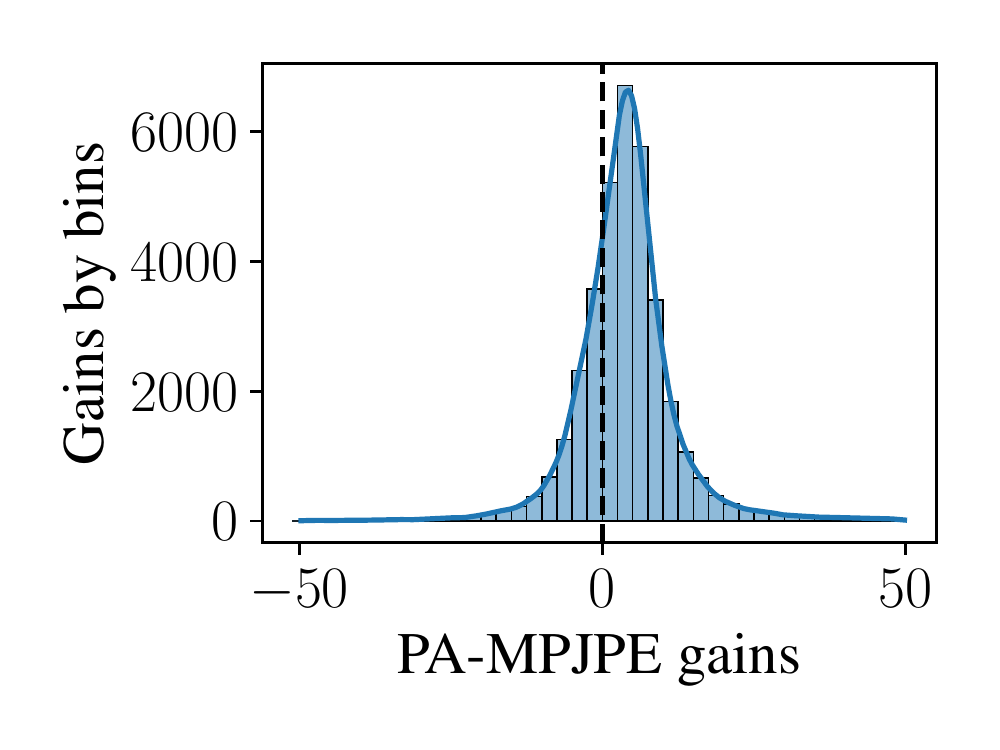}~
\includegraphics[width=0.5\columnwidth]{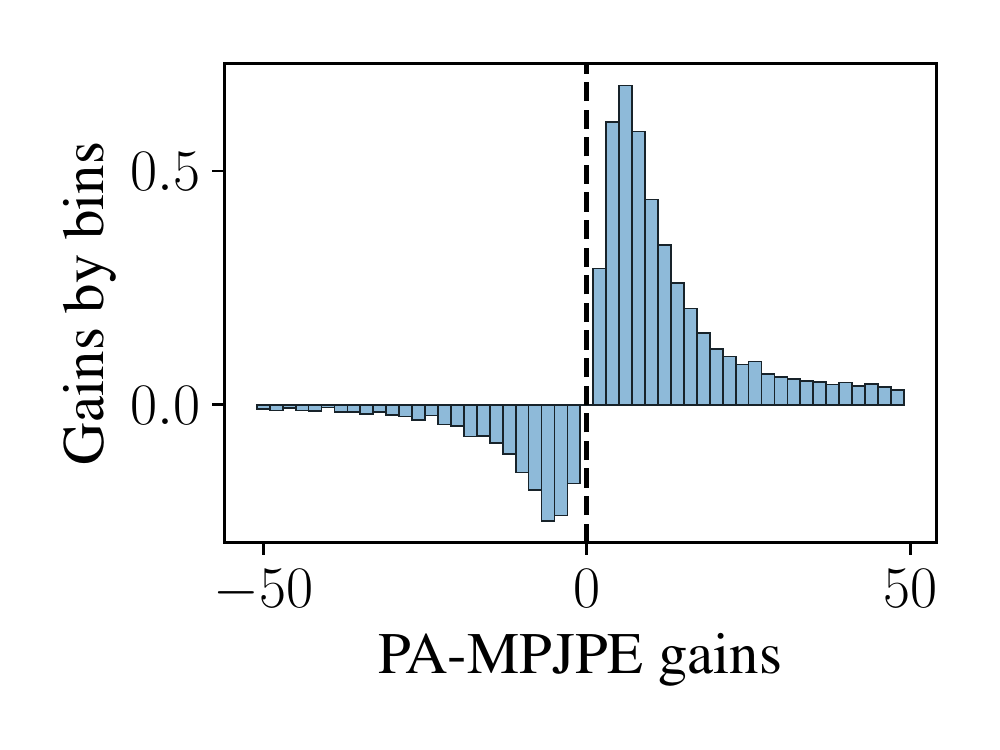} \\[-0.4cm]
\vspace{-0.15cm}
\caption{\label{fig:gain}\textbf{Performance gain analysis on 3DPW.} We study the gain of our \sspin  over SPIN~\cite{spin}. \textbf{Left:} Histogram of the gains in PA-MPJPE. Our method outperforms SPIN on 26393 test samples out of 35515. \textbf{Right:} Resulting performance gains for each bin in (left), obtained by multiplying the absciss of each bin with the fraction of samples inside it. Notice that  the bins corresponding to small improvements have many samples, and are the foundation of our 4 points gained in PA-MPJPE over SPIN.}
\vspace{-0.3cm}
\end{figure}

\paragraph{Analysis of the performance gain.}
To better understand in which case our proposed \sspin improves performance, we plot in Figure~\ref{fig:gain} (left) an histogram of the gain compared to the original SPIN model on 3DPW in terms of MPJPE. We observe that most samples are improved by a few millimeters. The overall shape of the histogram has a shape similar to a Gaussian, meaning that some images have actually a higher error, and a few samples are significantly improved. The histogram on the right in Figure~\ref{fig:gain} shows the impact of each bin on the final MPJPE metric.

\paragraph{Study of the real/synthetic domain gap.} 
In order to study the impact of the domain gap between real and synthetic data, we consider the test set of 3DPW as an oracle and gradually make the renderings closer to this oracle.
Sampling the SMPL pose parameters from 3DPW test set has low impact on the performance (0.3mm in MPJPE), meaning that the AMASS dataset sufficiently covers the pose space.
We then additionally replace the randomized choice of the global orientation by the ones of the corresponding relative pose from the 3DPW test set, thus rendering people in the same 3D global pose, but at random 2D positions in the image. This has an impact of 1 to 2mm, similar to the gain obtained when also keeping the exact same 2D positions by sampling the ground-truth camera parameters with the global pose. We finally keep the background image from the original data, in which case only the texture over the SMPL ground-truth is synthetically changed compared to the original images, and get an additional 1mm error reduction with a MPJPE of 85.2mm. 
Overall, \sspin is close in performance to the strongest oracle because training only the batch-normalization layers of the feature extraction network $f$ strongly constrains the learning problem.

\begin{table}
\centering
\resizebox{\linewidth}{!}{
\begin{tabular}{cccc|ccc}
\toprule
      GT relative & GT global       & GT camera       & \multirow{2}{*}{GT background} & \multicolumn{3}{c}{3DPW~\cite{3dpw}} \\
       pose    & orientation  &  translation &  & MPJPE $\downarrow$ & PA-MPJPE $\downarrow$ & MPVPE $\downarrow$ \\
\midrule
     & &  &  & 90.8 & 55.6 & 105.0\\
    \checkmark & &  &  & 90.5 & 54.1 & 104.8\\
    \checkmark & \checkmark &  &  & 88.0 & 53.1 & 102.3\\
    \checkmark & \checkmark & \checkmark &  & 86.1 & 52.9 & 100.5 \\
     \checkmark & \checkmark & \checkmark & \checkmark & \textbf{85.2} & \textbf{52.4} &  \textbf{99.7} \\
\bottomrule
\end{tabular}
}

\vspace{-0.3cm}

\caption{
\textbf{Oracle study on 3DPW.} Starting from \sspin (first row), we gradually add characteristics from the ground-truth (GT) 3DPW \textit{test} set
to the synthetic renderings, to make it closer to the real test data.
}
\label{tab:oracle_study}
\vspace{-0.6cm}
\end{table}

\begin{table}
\centering
\resizebox{\linewidth}{!}{
\begin{tabular}{lccc}
\toprule
 & 3DPW & MPI-INF-3DHP & MuPoTS-3D \\
\midrule
\sspin & 55.6 & 66.7	& 81.0 \\ 
\midrule
\sspin + \bf{\posebert} & \bf{53.2} & \bf{63.8} & \bf{79.9} \\
~~~~w/o pos. encoding & 54.8 & 64.0 & 80.8 \\
~~~~w/o shared regressor & 54.0 & 65.0 & 81.0 \\
~~~~with 2 regressor iterations & 53.3 & 64.0 & 80.5 \\
~~~~with 4 regressor iterations & 53.4 & 64.2 & 80.5 \\
\midrule
~~~~L=1 & 54.5 &	67.0 &	81.2 \\
~~~~L=2 & 53.9 & 	65.5 & 	81.0 \\
~~~~L=4 (default) & \bf{53.2} & 63.8 & 79.9 \\
~~~~L=8 & 53.4 &	\bf{63.3} &	\bf{79.8} \\
\midrule 
~~~~$D_t$=128	& 55.6	& 69.4 &	82.0 \\
~~~~$D_t$=256	& 53.7	& 64.0 &	80.4 \\
~~~~$D_t$=512 (default)	& \bf{53.2} &	63.8	& \bf{79.9} \\
~~~~$D_t$=1024	& 53.4	& \bf{62.9} 	& \bf{79.9} \\
\midrule
~~~~T=8 & 53.9	& 65.2 &	\bf{79.8} \\
~~~~T=16 (default) & \bf{53.2}	& 63.8 &	79.9 \\
~~~~T=32 & 53.4 &	\bf{63.6} & 80.3 \\
~~~~T=64 & 53.3 & 63.7 & 80.4 \\
\bottomrule
\end{tabular}
} %

\vspace{-0.3cm}

\caption{\textbf{Ablation on the \posebert hyperparameters.} We study the impact of the positional encoding, sharing the regressor, the number of regressor iterations per layer (1 by default), the depth L of the network (4 by default), the number of channels ($D_t$=512 by default) and the length of the sequences (T=16 by default) with the PA-MPJPE metric on 3DPW, MPI-INF-3DHP and MuPoTS-3D when using \posebert on top of \sspin, with masking 12.5\% of the input sequences (2 frames with T=16 frames).}
\label{tab:posebert_ablation}
\vspace{-0.6cm}
\end{table}

\begin{table*}[!t]
	\centering
	\resizebox{\textwidth}{!}{%
		\begin{tabular}{ll|cccc|ccc|ccc|ccc}
		    
			\specialrule{.1em}{.05em}{.05em}

			&  & \multicolumn{4}{c}{3DPW~\cite{3dpw}} & \multicolumn{3}{c}{MPI-INF-3DHP~\cite{mpiinf}} & \multicolumn{3}{c}{MuPoTS-3D~\cite{mupots}}& \multicolumn{3}{c}{AIST~\cite{aist}} \\
			\cmidrule(lr){3-6} \cmidrule(lr){7-9} \cmidrule(lr){10-12} \cmidrule(lr){13-15}
			\multicolumn{2}{l|}{Method} & $\mathcal{E}$ $\downarrow$ & MPJPE $\downarrow$ & MPVPE $\downarrow$ & Accel $\downarrow$ & $\mathcal{E}$ $\downarrow$ & MPJPE $\downarrow$ & Accel $\downarrow$ & $\mathcal{E}$ $\downarrow$ & MPJPE $\downarrow$ & Accel $\downarrow$& $\mathcal{E}$ $\downarrow$ & MPJPE $\downarrow$ & Accel $\downarrow$ \\
			
			\midrule

			\parbox[t]{2mm}{\multirow{6}{*}{\rotatebox[origin=c]{90}{single image}}}  & HMR~\cite{hmr} & 76.7 & 130.0 & - & 37.4 & 89.8 & 124.2 & - & - & - & - & - & - & - \\
			& \cellcolor{Gray}GraphCMR~\cite{graphcmr} & \cellcolor{Gray}70.2 & \cellcolor{Gray}- & \cellcolor{Gray}- & \cellcolor{Gray}- & \cellcolor{Gray}- & \cellcolor{Gray}-& \cellcolor{Gray}- & \cellcolor{Gray}- & \cellcolor{Gray}- & \cellcolor{Gray}-& \cellcolor{Gray}- & \cellcolor{Gray}- & \cellcolor{Gray}- \\
			& SPIN~\cite{spin} &  59.2 & 96.9 & 116.4 & 29.8 & 67.5 & 105.2 & 30.3 & $83.0^{\dag}$ & $154.6^{\dag}$ & $24.6^{\dag}$ & $76.2^{\dag}$ & $\bf{126.2}^{\dag}$ & $55.9^{\dag}$ \\
			& \cellcolor{Gray}I2L-MeshNet~\cite{i2lmeshnet} & \cellcolor{Gray}57.7 & \cellcolor{Gray}93.2 & \cellcolor{Gray} 110.1 & \cellcolor{Gray}30.9 & \cellcolor{Gray}- & \cellcolor{Gray}- & \cellcolor{Gray}- & \cellcolor{Gray}- & \cellcolor{Gray}- & \cellcolor{Gray}-& \cellcolor{Gray}- & \cellcolor{Gray}- & \cellcolor{Gray}- \\
			& Pose2Mesh~\cite{pose2mesh} & 58.3 & \textbf{88.9} & {106.3} & \textbf{22.6} & - & - & - & - & - & - & - & - & -\\
			&\cellcolor{Gray}\textbf{(Ours) \sspin} &\cellcolor{Gray} \textbf{55.6} &\cellcolor{Gray} 90.8 &\cellcolor{Gray} \textbf{105.0} & \cellcolor{Gray}27.5 &\cellcolor{Gray}   \textbf{66.7} &\cellcolor{Gray} \textbf{100.8} &\cellcolor{Gray} \textbf{29.5} &\cellcolor{Gray}  \textbf{81.0} &\cellcolor{Gray} \textbf{152.3} &\cellcolor{Gray} \textbf{24.2} & \cellcolor{Gray}  \textbf{75.7} & \cellcolor{Gray}125.1 &\cellcolor{Gray} \bf{55.7} \\
			\midrule

		    \parbox[t]{2mm}{\multirow{7}{*}{\rotatebox[origin=c]{90}{video}}} 
		    & HMMR~\cite{hmmr} & 72.6 & 116.5 & 139.3 & 15.2 &- & - & - & - & - & -& - & - & - \\
			& \cellcolor{Gray}Sun~\etal~\cite{sun2019human} & \cellcolor{Gray}69.5 & \cellcolor{Gray}- & \cellcolor{Gray}- & \cellcolor{Gray}- & \cellcolor{Gray}- & \cellcolor{Gray}- & \cellcolor{Gray}- & \cellcolor{Gray} -& \cellcolor{Gray}- & \cellcolor{Gray}-& \cellcolor{Gray}- & \cellcolor{Gray}- & \cellcolor{Gray}- \\
			&  VIBE~\cite{vibe} & 56.5 & 93.5 & 113.4 & 27.1 & 63.4 & 97.7 & 29.0 & 83.4$^{\dag}$ & 154.1$^{\dag}$ & 23.8$^{\dag}$ &  76.0$^{\dag}$ & 129.6$^{\dag}$ & 52.7$^{\dag}$ \\ %
			& \cellcolor{Gray}TCMR~\cite{tcmr} & \cellcolor{Gray}55.8 & \cellcolor{Gray}95.0 & \cellcolor{Gray}111.5 & \cellcolor{Gray}\textbf{7.0} & \cellcolor{Gray}\textbf{62.8} & \cellcolor{Gray}97.4 & \cellcolor{Gray}\textbf{8.0} & \cellcolor{Gray}82.9$^{\dag}$ & \cellcolor{Gray}152.8$^{\dag}$ & \cellcolor{Gray}\textbf{12.9}$^{\dag}$ & \cellcolor{Gray}75.3$^{\dag}$ & \cellcolor{Gray}126.9$^{\dag}$ & \cellcolor{Gray}\textbf{13.5}$^{\dag}$ %
			\\ 
            & \textbf{(Ours) \sspin + \posebert} & \textbf{52.9} & 89.4 & \textbf{103.8} & 8.3 &   63.3 & 97.4 & 8.7 &  \textbf{79.9} & \textbf{151.1} & 13.7 &   \textbf{74.1} &  \textbf{123.8} &  14.0
            \\
            \specialrule{.1em}{.05em}{.05em} \\
		\end{tabular}%
	}

\vspace{-0.5cm}
	
\caption{
\textbf{Comparison with state-of-the-art methods} on 3DPW~\cite{3dpw}, MPI-INF-3DHP~\cite{mpiinf}, MuPoTS-3D~\cite{mupots} and AIST~\cite{aist}. For brevity, we denote the PA-MPJPE metric as  $\mathcal{E}$. Methods are organized by input type (single image and video). All methods listed follow the standard protocol and do not use the 3DPW training data. Unless otherwise stated, results are copied from the corresponding papers. The symbol ${\dag}$ denotes results obtained by running the model released by the authors.
}
	
\label{tab:sota}
\vspace{-0.5cm}
\end{table*}

\begin{table}
\centering
\resizebox{\linewidth}{!}{
\begin{tabular}{lcc|cc|cc}
\toprule
 & \multicolumn{2}{c|}{3DPW} & \multicolumn{2}{c|}{MPI-INF-3DHP} & \multicolumn{2}{c}{MuPoTS-3D} \\
 Masking \% & $\mathcal{E}$ $\downarrow$ & Accel $\downarrow$ & $\mathcal{E}$ $\downarrow$ & Accel $\downarrow$ & $\mathcal{E}$ $\downarrow$ & Accel $\downarrow$ \\
\midrule
\sspin & 55.6 & 32.5 &	66.7 & 29.5 &  81.0 & 23.5 \\
\midrule 
 0\% &    53.3 & 9.6 & \bf{62.3} & 9.8 & 80.0 & 13.8 \\
12.5\% & 53.2 & \bf{7.8}  & 63.8 & \bf{8.7} & 80.3 & \bf{12.8} \\
25\% &   53.3 & 8.3 & 64.2 & 9.0 & 80.3 & 13.3 \\
37.5\% & 53.9 & 9.0 & 65.0 & 9.2 & 80.8 & 14.0 \\
\midrule
12.5\% + Noise & \bf{52.9} & 8.3 & 63.3 & \bf{8.7} & \bf{79.9} & 13.7 \\
\bottomrule
\end{tabular}
} %
\vspace{-0.3cm}
\caption{\textbf{Ablation on the \posebert pretraining strategy.} We study the impact of masking the input sequences and adding Gaussian noise. $\mathcal{E}$ refers to the PA-MPJPE metric.}
\label{tab:posebert_pretraining}
\vspace{-0.3cm}
\end{table}

\begin{table}
\centering
\resizebox{\linewidth}{!}{
\begin{tabular}{lllll}
\toprule
 & \multicolumn{1}{c}{3DPW} & \multicolumn{1}{c}{MPI-INF-3DHP} & \multicolumn{1}{c}{MuPoTS-3D} & \multicolumn{1}{c}{AIST} \\
\midrule 
SPIN~\cite{spin} & 59.6	& 68.0 & 83.0 & 76.2 \\
~~~~+ \posebert & \bf{57.3} \errorgain{2.3} & \bf{64.3} \errorgain{3.7} & \bf{80.9} \errorgain{2.1} & \bf{74.6} \errorgain{1.6} \\
\midrule
VIBE~\cite{vibe} & 56.5 & 65.4 & 83.4 & 76.0 \\
~~~~+ \posebert & \bf{54.9} \errorgain{1.6} & \bf{64.4} \errorgain{1.0} & \bf{81.0} \errorgain{2.4} & \bf{74.5} \errorgain{1.5} \\ 
\midrule
\sspin & 55.6 & 66.7 & 81.0 & 75.7 \\ 
~~~~+ \posebert & \bf{52.9} \errorgain{2.7} & \bf{63.8} \errorgain{2.9} & \bf{79.9}   \errorgain{1.1} & \bf{74.1} \errorgain{1.6}\\
\bottomrule
\end{tabular}
}

\vspace{-0.3cm}

\caption{\textbf{Adding \posebert on top of various methods.} We report the PA-MPJPE metric (lower is better) on four video datasets. The gains in mm are shown in parenthesis.}
\label{tab:posebert_impact}
\vspace{-0.6cm}
\end{table}

\subsection{Evaluation of \posebert}
\label{sub:xpposebert}

In this section, we evaluate and ablate \posebert, our video-based model trained using MoCap data only.

\paragraph{Study of different architectures.}
As a first step we study the impact of some architecture design choices and report the PA-MPJPE on three datasets in Table~\ref{tab:posebert_ablation}.
As a default training strategy we mask 12.5\% of the input poses for this ablation.
First we note that \posebert leads to a consistent gain of 1 to 3mm on all datasets.
Removing the positional encoding leads to a suboptimal performance indicating that incorporating temporal information within the network is a key design choice.
Sharing the regressor allows to reduce the number of learnable parameters and leads to better predictions.
More importantly we notice that we can iterate over the regressor only a single time after each layer given that doing more iterations does not improve and even slightly decreases the performance.
In terms of model size, the benefit of \posebert seems to be reached with a depth of $L=4$ and an embedding dimension of $D_t=512$.
We choose these hyperparameters since increasing the model complexity leads to minimal improvements.
For the temporal length of the training sequence, we set $T=16$, as longer sequences do not lead to further improvements.

\paragraph{Training strategies.}
We then study the impact of various training strategies on MoCap datasets in Table~\ref{tab:posebert_pretraining}.
First, we study the impact of partially masking the input sequences, and observe that masking 12.5\%, \ie, 2 frames out of 16, lead to smoother prediction (lower error acceleration) while the PA-MPJPE remains low.
We also try adding Gaussian noise, with a standard deviation of 0.05 on top of the axis-angle representation, and obtain a small additional boost of performance and smoother predictions.
Increasing the standard deviation did not bring any benefit.
Motivations for adding Gaussian noise is described in the Supplementary Material.

\paragraph{Plugging \posebert on top of existing approaches.}
One major key benefit of \posebert is that it can be plugged on top of any image-based model to transform it into a video-based model since it takes \textit{only} SMPL sequences as input compared to other methods (VIBE, TCMR) which require image-based features as input.
In Table~\ref{tab:posebert_impact}, we report the PA-MPJPE on the 4 video datasets. We observe that when plugging \posebert on top of SPIN, it leads to a consistent improvement of 2.3mm on 3DPW, 3.7mm on MPI-INF-3DHP, 2.1mm on MuPoTS-3D and 1.6mm on AIST.
Interestingly, this improvement is higher than the one obtained when using VIBE on MPI-INF-3DHP, MuPoTS-3D and AIST. When using \sspin as image-based model, we observe a similar consistent improvement on all datasets. Actually, one can even plug \posebert on top of a model that already leverages videos, such as VIBE~\cite{vibe}, and we observe a similar consistent gain, which suggests that \posebert is complementary to the way temporal consistency of features is exploited in VIBE.
Finally, to test it beyond SMPL inputs, we appended PoseBERT on top of 2D/3D pose predictions from LCRNet++ \cite{lcrnet++} and see significant gains on 3DPW, a decrease of MPJPE (resp. PA-MPJPE) from 125.8 (resp. 68.8) to 108.2 (resp. 58.5). More details are given in the supplementary materiel.

\subsection{Comparison to the state of the art}
\label{sub:xpsota}

Finally we compare our image-based and video-based methods against previous works in Table~\ref{tab:sota}.

\paragraph{Single image.} We compare \sspin with HMR~\cite{hmr}, GraphCMR~\cite{graphcmr}, SPIN~\cite{spin}, I2L-MeshNet~\cite{i2lmeshnet}, and Pose2Mesh~\cite{pose2mesh} on four datasets.
\sspin outperforms all other methods on all datasets in PA-MPJPE, for instance by 6.1mm MPJPE on 3DPW.

\paragraph{Video.} Finally we compare \posebert against other video-based methods \cite{vibe,hmmr}.
We emphasize that compared to other approaches which require image-based features, \posebert is trained only using MoCap data from AMASS and takes SMPL sequences as input.
For 3DPW \cite{3dpw}, \posebert outperforms concurrent works on the PA-MPJPE and  MPVPE metrics.
We do not see any improvement by finetuning on 3DPW train.
For MPI-INF-3DHP, \posebert demonstrates results on par with the state of the art while it is worth mentioning that, compared to all other methods, our temporal module does not use the associated training set.
Finally on MuPoTS-3D~\cite{mupots} and AIST~\cite{aist}, \posebert shows gains on all metrics over VIBE and TCMR.

%% file: tex/05_conclusions.tex
\section{Conclusion}
\label{sec:conclusion}

In this paper we propose two ways of leveraging MoCap data to improve image and video-based human 3D mesh recovery. We first present  \sspin, a very strong baseline that reaches state-of-the-art performance for image-based models on a number of datasets that is obtained by simply fine-tuning the batch normalization layer parameters of SPIN with the use of real data and  synthetic renderings of MoCap data.  We further introduce \posebert, a transformer module that directly regresses the SMPL pose parameters and is purely trained on MoCap data via masked modeling. Our experiments show that \posebert can be readily plugged on top of any image-based model to leverage temporal context and improve its performance.

%% file: tex/06_supp_mat.tex
\newpage
\section{Appendix}
\label{sec:supp_mat}

In this supplementary material, we first present additional results and illustrations for \sspin (Section~\ref{app:sspin}) and \posebert (Section~\ref{app:posebert}).
We then present Figure~\ref{fig:extended_teaser} an extended version of paper's Figure 1, depicting qualitative results for additional pose estimation methods.

\section{Additional results for \sspin}
\label{app:sspin}

In this section, we first illustrate our oracle experiments presented in Table 2 of the main paper as well as an interesting experiment where we evaluate SPIN on synthetic renderings (Section~\ref{appsub:oracle}).
We then present more results on the fine-tuning of \sspin (Section~\ref{appsub:sspin_finetuning}). In particular, we ablate the choice of only fine-tuning the batch-normalization layers.
We finally study the impact of using feature alignment loss when using paired data (Section~\ref{appsub:alignment}).

\begin{figure*}
\centering
\small
{
\newlength{\oraclefigwidth}
\setlength{\oraclefigwidth}{0.15\linewidth}
\setlength{\tabcolsep}{1pt}
\begin{tabular}{cccccc}
Real test image & \multicolumn{4}{c}{$\xrightarrow{\hspace*{4\oraclefigwidth}}$} & AMASS rendering \\
\multicolumn{1}{c}{\includegraphics[width=\oraclefigwidth]{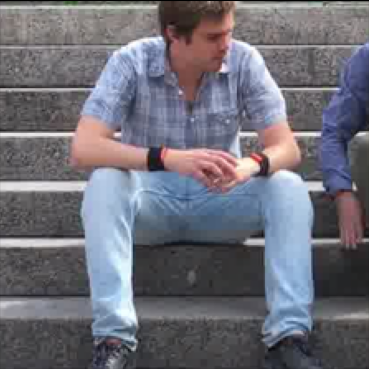}}
& \includegraphics[width=\oraclefigwidth]{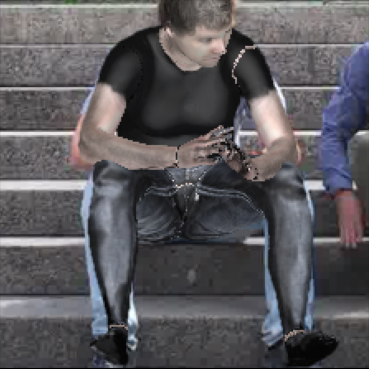}
& \includegraphics[width=\oraclefigwidth]{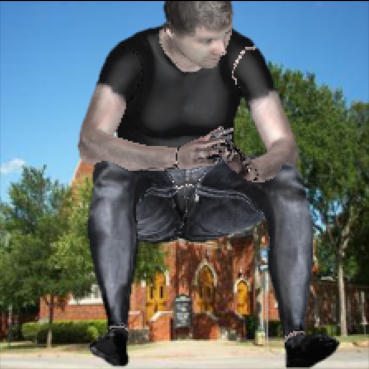}
& \includegraphics[width=\oraclefigwidth]{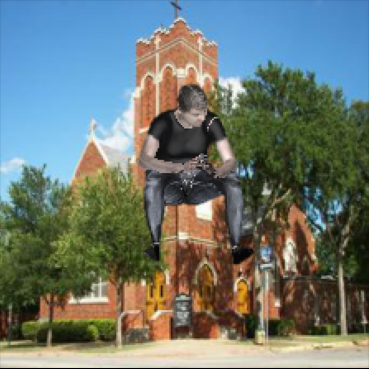}
& \includegraphics[width=\oraclefigwidth]{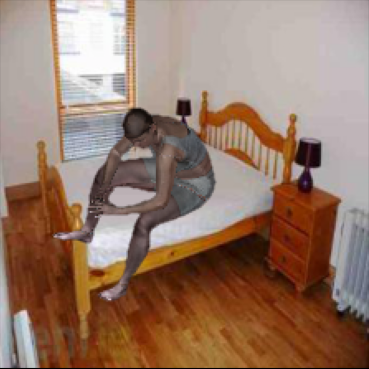}
& \includegraphics[width=\oraclefigwidth]{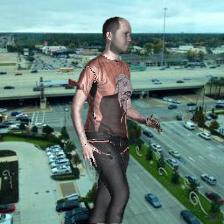}\\

    \toprule
GT relative pose      & \checkmark & \checkmark & \checkmark &  \checkmark \\
\midrule
GT global orientation & \checkmark & \checkmark & \checkmark & \\
\midrule
GT camera translation & \checkmark & \checkmark &            & \\
\midrule
GT background         & \checkmark &            &            & \\
\bottomrule
\end{tabular}
}
\caption{\label{fig:oracle} \textbf{Data samples used in oracle experiments.}
Left: real image from 3DPW~\cite{3dpw} test set.
Right: synthetic rendering of a pose from AMASS~\cite{amass} MoCap dataset.
Middle: synthetic renderings exploiting parts of ground truth annotations, for oracle experiments.
}
\end{figure*}

\subsection{Real to synthetic renderings domain gap}
\label{appsub:oracle}

\paragraph{Illustration of the oracle experiment renderings.}
Figure~\ref{fig:oracle} illustrates the different kind of renderings used in our oracle experiments (Table 2 of the main paper).
Starting from a real test image (left), we gradually remove some parts of the ground-truth real data, \ie, background, camera translation, global orientaiton and pose. The latter case is \sspin's setting, where we do not use any test data information, and only use renderings from the AMASS dataset.

\begin{figure*}
\centering
\small
{
\newlength{\spinfigwidth}
\setlength{\spinfigwidth}{0.17\linewidth}
\setlength{\tabcolsep}{1pt}
\begin{tabular}{c|ccccc}
& \includegraphics[width=\spinfigwidth]{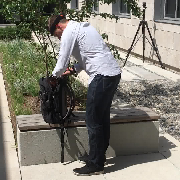}
& \includegraphics[width=\spinfigwidth]{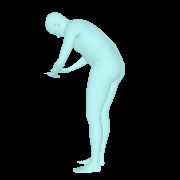}
& \includegraphics[width=\spinfigwidth]{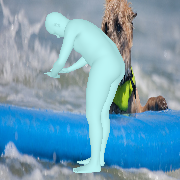}
& \includegraphics[width=\spinfigwidth]{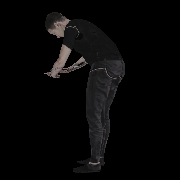}
& \includegraphics[width=\spinfigwidth]{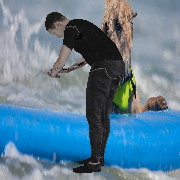}\\

    \toprule
Background      & Real image & Black & Random Coco image &  Black &  Random Coco image \\
Texture & Real image & Metallic & Metallic & SURREAL texture & SURREAL texture  \\
\midrule
MPJPE$\downarrow$ & 97.2 & 117.7 & 135.1 & {\bf 93.5} & 103.5          \\
PA-MPJPE$\downarrow$ & 59.6 & 63.8 & 72.8 & {\bf 56.4} & 64.1                      \\
MPVPE$\downarrow$  & 116.8 & 137.0 & 157.2 & {\bf 112.3} & 125.3                    \\
\bottomrule
\end{tabular}
}
\caption{\label{fig:spin} \textbf{Using SPIN pretrained model to study the domain gap between real and synthetic renderings on 3DPW.} We run SPIN on various synthetic renderings of 3DPW~\cite{3dpw} using different backgrounds and textures. SPIN generalizes to renderings with random backgrounds and SURREAL textures. }%
\end{figure*}

\paragraph{SPIN on synthetics.}
 The gap between in-the-wild images and renderings of MoCAP data is central to our fine-tuning strategy. We aim to learn pose estimation from synthetic renderings and generalize to in-the-wild image.  To better understand the domain gap, we study an opposite but informative problem : how does a model trained on in-the-wild images generalize to synthetic renderings? To that end, we generated several versions of 3DPW. For each data sample of 3DPW, we render the ground-truth pose using the same camera parameters, using either black or random background, and either metallic or SURREAL textures. We compare the performance of SPIN on each dataset version in Figure~\ref{fig:spin}. We observe two phenomenons. First, SPIN performs better on black backgrounds than random backgrounds. Segmenting the human in the crop does make the prediction task easier. Second, SPIN performs better when SURREAL textures are used instead of metallic textures. Indeed, SURREAL textures makes the renderings look more realistic and reduces the domain gap. Most importantly, the strong performance of SPIN on renderings using random backgrounds and SURREAL textures (-4.5mm of PA-MPJPE) motivates \sspin. Indeed, we empirically observe that SPIN, a model trained on in-the-wild images, can generalize to synthetic data. This means that the the domain gap between in-the-wild and synthetic data is not huge. In the main paper, we show that \sspin (our fine-tuning that uses additional synthetic renderings) brings a 4mm improvement in PA-MPJPE over SPIN on the real-image version of 3DPW.

\subsection{Training different parts of the \sspin network.}
\label{appsub:sspin_finetuning}

\begin{figure}
\centering
\newlength{\overfittingfigwidth}
\setlength{\overfittingfigwidth}{0.95\linewidth}
    \includegraphics[width=\overfittingfigwidth]{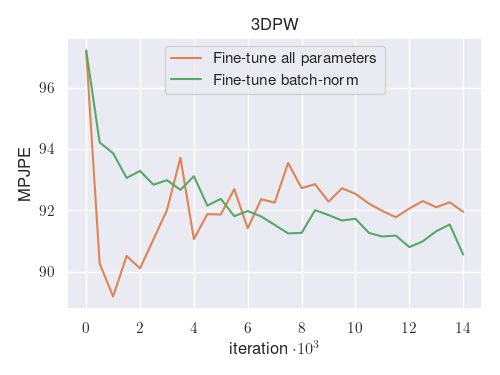}\\
    \includegraphics[width=\overfittingfigwidth]{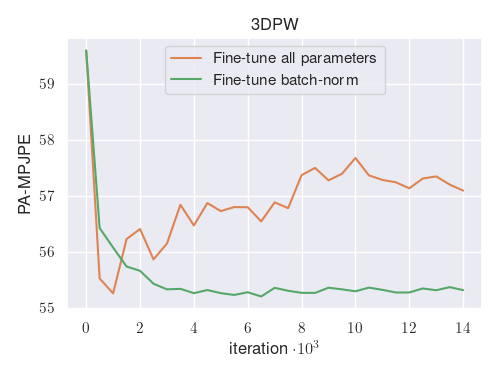}
\caption{\label{fig:overfitting} \textbf{Fine-tuning all parameters lead to overfitting.} We compare fine-tuning  all the parameters of SPIN against only the batch-normalization layers. Fine-tuning  all  parameters leads to a clear case of overfitting. On 3DPW, the best performance is reached after 500 iterations, with batches of size 64. On the other hand, fine-tuning only the batch-norm parameters (\sspin) leads to monotonically decreasing errors.  We report the MPJPE {\bf(top)} and PA-MPJPE {\bf(bottom)} on 3DPW test set.
}
\end{figure}

In this section, we study the impact of fine-tuning only the batch-normalization layers of \sspin. In Table~\ref{tab:finetuning_parameters}, we study the performance of fine-tuning different groups of parameters of SPIN. Fine-tuning only the batch-normalization layers outperforms other settings. In particular, fine-tuning all the parameters leads to overfitting, which we show on 3DPW in Figure~\ref{fig:overfitting} and is the central motivation to only fine-tuning batch-normalization layers.

\begin{table}[h]
\centering
\resizebox{\linewidth}{!}{
\begin{tabular}{lccccc}
\toprule
 & \% trainable params. & 3DPW & MPI-INF-3DHP & MuPoTS-3D & AIST\\
\midrule
SPIN & - & 59.6 & 68.0	& 83.0 & 71.8 \\ 
\midrule
All parameters & 100 & 57.1 & 66.7 & 81.7 & 71.8 \\
Regressor & 12.8 & 56.6 & 67.5 & 82.5 & 71.9 \\
Regressor + last layer of feature extractor &  68.3 & 56.5 & 66.9 & 
81.3 & 71.6 \\
Regressor + last two layers of feature extractor & 94.6 & 57.3  & 66.9 & 81.7 & 71.6 \\
Regressor + last three layers of feature extractor & 99.2 &  57.0 & 66.7 & 81.4 & 72.0 \\
Regressor + last four layers of feature extractor & 99.9 & 57.5  & 67.0 & 81.8 & 71.9 \\
\midrule
All but batch-norm & 99.8 & 57.6 & 66.9 & 81.8 & 71.6\\
Batch-norm + regressor & 13.0 & \bf{55.4} & \bf{66.5} & 81.6 & \bf{71.2}\\
Batch-norm (\sspin) & 0.2 & 55.6 & 66.7 & \bf{81.0} & 71.6 \\

\bottomrule
\end{tabular}
} 
\caption{\textbf{Additional ablation on  \sspin trainable parameters.} We compare fine-tuning only the batch-normalization layers~\cite{frankle2020training} with other subsets of SPIN parameters. }
\label{tab:finetuning_parameters}
\end{table}

\subsection{Feature alignment loss on paired data}
\label{appsub:alignment}

 SPIN  is pretrained on a set of datasets that contain 2D and 3D pose ground-truth data~\cite{lsp,mpii,coco}. \sspin leverages additional poses from Amass~\cite{amass}. We compare  in Table 2 of the main paper adding AMASS data against rendering poses from the original SPIN datasets. In the latter case,  we utilize the pseudo ground-truth pose information of image $\vx$ to render an image $\vxsynth$ of a synthetic human in the same pose. In practice, we use the final fits from the SPIN model~\cite{spin} as pseudo ground-truth SMPL parameters for these real-world datasets.
For every pair $\vx, \vxsynth$, we extract feature vectors $\phi_\vx, \phi_\vxsynth$.  Following the domain adaptation literature, we argue that maximizing the similarity between the two features can lead to better domain invariance. The benefits of the latter are twofold: (a) features from synthetic images are better aligned with features from real images, therefore synthetic renderings can be used for learning and (b)  features should be more robust to background appearance changes. 

We therefore use a feature alignment loss $L_{\text{align}} = - \lambda s(\phi_\vx, \phi_\vxsynth)$ for every pair $\vx, \vxsynth$, where $s(\phi_\vx, \phi_\vxsynth)$ is a similarity function between the two feature vectors that can be defined in a number of ways. 
For example, one can define $s$ as the negative of the mean square error or L2 distance, the cosine similarity, or the recently popular contrastive loss~\cite{simclr}.

In practice,  we use :
\begin{equation}
\begin{split}
s_{\text{MSE}}(\phi_\vx, \phi_\vxsynth)  & = - || \phi_\vx - \phi_\vxsynth ||^2_2 \\
s_{\text{cosine}}(\phi_\vx, \phi_\vxsynth)  & =  \dfrac{\phi_\vx \cdot \phi_\vxsynth}{\max(\Vert \phi_\vx \Vert _2 \cdot \Vert \phi_\vxsynth \Vert _2, \epsilon)} \\
s_{\text{contr}}(\phi_\vx, \phi_\vxsynth) & = - \log \frac{\exp(\phi_\vx^T \phi_\vxsynth)}{\sum_{\vy \in \mathcal{N}} \exp(\phi^T_{\vx} \phi_\vy)} ,
\end{split}
\end{equation}
where $\mathcal{N}$ is the set of all synthetic images in the batch. For the contrastive loss case, we first $\ell2$-2 normalize the features and  symmetrise the loss by  setting $s(\phi_\vx, \phi_\vxsynth) = s_{\text{contr}}(\phi_\vx, \phi_\vxsynth) + s_{\text{contr}}( \phi_\vxsynth, \phi_\vx)$. The contrastive loss is essentially a softmax function bringing the correct real-synth features closer to each other and further from all other features from synthetic images in the same batch.

We report results in Table~\ref{tab:single_image_feature_loss}. For each loss type, we conduct an hyperparameter search to select the best weight $\lambda$. Using the L2 or cosine losses to align the feature do not yield an improvement. We view this as a valuable negative result. Aligning features with the contrastive loss yields a clear quantitative improvement on 3DPW, MPI-INF-3D and MCB and outperforms \sspin by 4.2mm of MPJPE on AIST. We explored combining this experiment with \sspin \ie using additional data from amass \textit{and} aligning paired samples with a contrastive loss, but  did not observe performance gains compared to \sspin. 

\begin{table*}[t]
\centering
\resizebox{\linewidth}{!}{
\begin{tabular}{lccc|cc|cc|cc|c}
\toprule
      & \multicolumn{3}{c|}{3DPW~\cite{3dpw}} & \multicolumn{2}{c|}{ MuPoTS-3D~\cite{mupots} } & \multicolumn{2}{c|}{ AIST~\cite{aist} } & \multicolumn{2}{c|}{ MPI-INF-3D~\cite{mpiinf} } & MCB~\cite{mc3dv} \\
      & MPJPE $\downarrow$ & $\mathcal{E}$ $\downarrow$ & MPVPE $\downarrow$ & MPJPE $\downarrow$ & $\mathcal{E}$ $\downarrow$ & MPJPE $\downarrow$ & $\mathcal{E}$ $\downarrow$ & MPJPE $\downarrow$ & $\mathcal{E}$ $\downarrow$ & MPJPE $\downarrow$ \\
    \midrule
     \textbf{SPIN~\cite{spin}} & 97.2 & 59.6 & 116.8 & 154.6 & 83.0 & 119.4 & 71.8 & 104.3 & 68.0  & 155.4 \\
    \midrule
     \textbf{Using paired renderings of SPIN data} & & & & & \\
    \textit{~~~~~~ + pose losses on rendered images} & 93.5 & 58.6 & 109.7 &{152.2} & {\bf 82.1} & 117.4 & {\bf 71.8} & 98.0 & 67.4 & 150.0 \\
    \textit{~~~~~~~~~~~~  +  L2 loss}  & 93.4  & 58.7  & 109.3 & {\bf 152.1} & {\bf 82.1} & 119.2 & 72.7 & 98.5 & 67.4 & 149.5 \\    
    \textit{~~~~~~~~~~~~  +  cosine loss}  & 93.5 & 58.7  & 109.6 & 152.4 & 82.2 & 119.3 & 72.8 & 98.4 & 67.5 & 149.9 \\    
    \textit{~~~~~~~~~~~~  +  contrastive loss}  & {\bf 92.7}  & {\bf 58.0}  & {\bf 106.1} & 152.5 & {\bf 82.1} & \textbf{116.6} & 72.5 & \textbf{97.3} & {\bf 66.8} & {\bf 146.7}\\    \midrule
     \textbf{Using renderings of MoCap data} & & & & & \\
     \sspin & \textbf{90.8} & \textbf{55.6} & \textbf{105.0} & 152.3 & \textbf{81.0} & 120.8 &  {71.6} & 100.8 & 66.7 & \textbf{145.0} \\
\bottomrule
\end{tabular}
}
\caption{
{
\textbf{Additional experiments with paired real/synthetic data.}  For brevity, we denote the PA-MPJPE metric as  $\mathcal{E}$. We experiment with different losses to align features between a real image and its synthetic rendering using a pseudo ground-truth. We conducted a hyperparameter search for each type of feature alignment loss. We weight by $10^{-3}$  the L2 and cosine loss, and by $10^{-2}$ the contrastive loss.}
}
\label{tab:single_image_feature_loss}
\end{table*}

\subsection{Variance of \sspin performance.}

We fine-tune \sspin 10 times with different random seeds and report the result in Table~\ref{tab:Mocap_spin_seed}. We observe that our fine-tuning strategy gives consistent and stable improvements over SPIN in each case.

\begin{table}
\centering
\resizebox{\linewidth}{!}{
\begin{tabular}{lccc|cc|cc|cc}
\toprule
      & \multicolumn{3}{c|}{3DPW~\cite{3dpw}} & \multicolumn{2}{c|}{ MuPoTS-3D~\cite{mupots} } & \multicolumn{2}{c|}{ AIST~\cite{aist} } & \multicolumn{2}{c|}{ MPI-INF-3D~\cite{mpiinf} }  \\
      \sspin & MPJPE $\downarrow$ & $\mathcal{E}$ $\downarrow$ & MPVPE $\downarrow$ & MPJPE $\downarrow$ & $\mathcal{E}$ $\downarrow$ & MPJPE $\downarrow$ & $\mathcal{E}$ $\downarrow$ & MPJPE $\downarrow$ & $\mathcal{E}$ $\downarrow$  \\
    \midrule
Mean & 91.1 & 55.4 & 105.2 & 152.2 & 81.0 & 120.9 & 71.5 & 100.5 & 66.5 \\
Min & 90.2 & 55.2 & 104.3 & 151.8 & 80.8 & 120.2 & 71.3 & 100.0 & 66.4 \\
Max & 92.2 & 55.5 & 106.2 & 152.6 & 81.2 & 121.3 & 71.6 & 100.9 & 66.7 \\
Median & 91.1 & 55.4 & 105.1 & 152.2 & 81.0 & 120.9 & 71.4 & 100.5 & 66.5 \\
Std & 0.68 & 0.08 & 0.67 & 0.21 & 0.11 & 0.36 & 0.09 & 0.26 & 0.07 \\
\bottomrule
\end{tabular}
}
\caption{
{
\textbf{Variance of \sspin}. We fine-tune SPIN with 10 difference random seed and report the mean, min , max, median and standard deviation of each metric. We conclude that our experiments are stable.}
}
\label{tab:Mocap_spin_seed}
\end{table}

\subsection{Implementation details for \sspin}
\label{appsub:image_details}

Our codebase is based on the official Pytorch~\cite{pytorch} SPIN release: \url{https://github.com/nkolot/SPIN}. In particular, we use the same set of pose losses with the same weights.
For \sspin, in addition to updating the $5.3 \cdot 10^4$ running statistics, we train the $5.3 \cdot 10^4$ parameters of the batch-norm affine layers out of the $2.7 \cdot 10^7$ SPIN trainable parameters (0.2\%).
We fine-tune for 14k iterations using the Adam optimizer~\cite{adam}, with batches of size 64, with a learning rate of $3 \cdot 10^{-5}$, divided by 10 after 10k and 12k iterations. Training takes about 10 hours on a NVIDIA V100 graphics card. The speed bottleneck is on the CPU side and speed could be improved with more dataloader workers.

\section{Additional results for \posebert}
\label{app:posebert}

In this section, we first study the impact on finetuning \posebert on real-world training data (Section~\ref{appsub:finetune}) as well as the impact of other architectures and training strategies (Section~\ref{appsub:training}).

\subsection{Fine-tuning \posebert on real-world data}
\label{appsub:finetune}

While we propose to train \posebert from scratch using MoCap data only, we also study the impact of fine-tuning on real-world training sets from several datasets, see results in Table~\ref{tab:posebert_finetuning}.
We observe that finetuning leads to a boost on the associated test sets but may decrease performances on other test sets.
For example, finetuning on the MPI-INF-3DPH training set improves the results on MPI-INF-3DPH and MuPoTs-3DPH respectively by 3.3 and 0.5 mm but it decreases the performance of \posebert on 3DPW by 1 mm.
This can be explained by the domain shift between these specific training sets.

\begin{table}[h]
\centering
\resizebox{\linewidth}{!}{
\begin{tabular}{lcccc}
\toprule
 & 3DPW & MPI-INF-3DHP & MuPoTS-3D & AIST \\
\midrule
\sspin & 55.6 & 66.7	& 81.0 & 71.6 \\ 
\midrule
\sspin + \bf{\posebert} & 53.2 & 63.8 & 80.3 & 69.7 \\
~~~~~~~~ \textit{+ finetuning on MPI-INF-3DPH} & 53.9 &  \bf{60.5} & \bf{79.8} & 71.5 \\
~~~~~~~~ \textit{+ finetuning on 3DPW} & \bf{52.9} &  63.8 & 79.9 & 70.9  \\
~~~~~~~~ \textit{+ finetuning on AIST} & 53.9  & 65.6 & 80.9 & \bf{69.5} \\
\bottomrule
\end{tabular}
}
\caption{\textbf{Fine-tuning on real data.} We study the impact of fine-tuning \posebert on different training sets. We reported the PA-MPJPE on different datasets.}
\label{tab:posebert_finetuning}
\end{table}

\subsection{\posebert: other training strategies}
\label{appsub:training}

In addition to masking and adding Gaussian noise on the input, we have also investigated other training strategies as reported in Table~\ref{tab:posebert_ablation_bis}.
First we compare against the common practice of having the iterative regressor \cite{thmmr} on top of the temporal module.
\posebert shows a gain ranging from 1.4 mm to 0.4mm compare to the baseline described above.
Then we increase the temporal window of the input sequence by reducing the frames per second while keeping the sequence length fixed.
We observe that increasing the time span does not bring significant improvement and even leads to decrease performances.
We also study the impact of incorporating random poses or joints compared to random Gaussian noise as proposed in the main paper.
We note that both noise types bring a small improvement compared to Gaussian noise but for simplicity we do not include them during the training scheme of our best model.

\begin{table}[h]
\centering
\resizebox{\linewidth}{!}{
\begin{tabular}{lccc}
\toprule
 & 3DPW & MPI-INF-3DHP & MuPoTS-3D \\
\midrule
\sspin & 55.6 & 66.7	& 81.0 \\ 
\midrule
\sspin + Transformer + Regressor & 54.5 & 65.2 & 80.7 \\
\midrule
\sspin + \bf{\posebert} & \bf{53.2} & 63.8 & \bf{80.3} \\
~~~~fps=7.5 & 54.0 & 64.5 & 80.5 \\
~~~~fps=15 & 53.4 & \bf{63.4} & 80.4 \\
~~~~fps=3.75 & 55.0 & 66.0 & 81.0 \\
\midrule
~~~~Random poses ~~5\% & 53.2 & \bf{63.5} & \bf{80.0} \\
~~~~~~~~~~~~~~~~~~~~~~~~~~~~ 10\%  & \bf{53.2} & 63.7 & \bf{80.0} \\
~~~~~~~~~~~~~~~~~~~~~~~~~~~~ 20\% & \bf{53.2} & 64.3 &  80.2 \\
~~~~~~~~~~~~~~~~~~~~~~~~~~~~ 40\%  & 53.6 & 64.5 &  80.4\\
\midrule 
~~~~Random joints ~~5\% & \bf{53.2} & \bf{62.7} & \bf{80.1} \\
~~~~~~~~~~~~~~~~~~~~~~~~~~~~ 10\%  & 53.4 & 63.6 & 80.2 \\
~~~~~~~~~~~~~~~~~~~~~~~~~~~~ 20\% & 53.3 & 64.9 & 81.0 \\
~~~~~~~~~~~~~~~~~~~~~~~~~~~~ 40\%  & 55.8 & 70.0 & 83.0 \\
\bottomrule
\end{tabular}
} %
\caption{\textbf{Additional ablation on the \posebert hyperparameters.} We first study the impact of having the regressor incorporated into the transformer. We also study the impact of the frame per second (fps, 30 by default) and the percentage of random poses/joints (0 by default) with the PA-MPJPE metric on 3DPW, MPI-INF-3DHP and MuPoTS-3D when using \posebert on top of \sspin, with masking 12.5\% of the input sequences (2 frames with T=16 frames) and using a model of size $D=512$ and $L=4$. 
}
\label{tab:posebert_ablation_bis}
\end{table}

\subsection{Motivation and details on adding Gaussian noise}
\begin{figure}
    \centering
    \includegraphics[width=\linewidth]{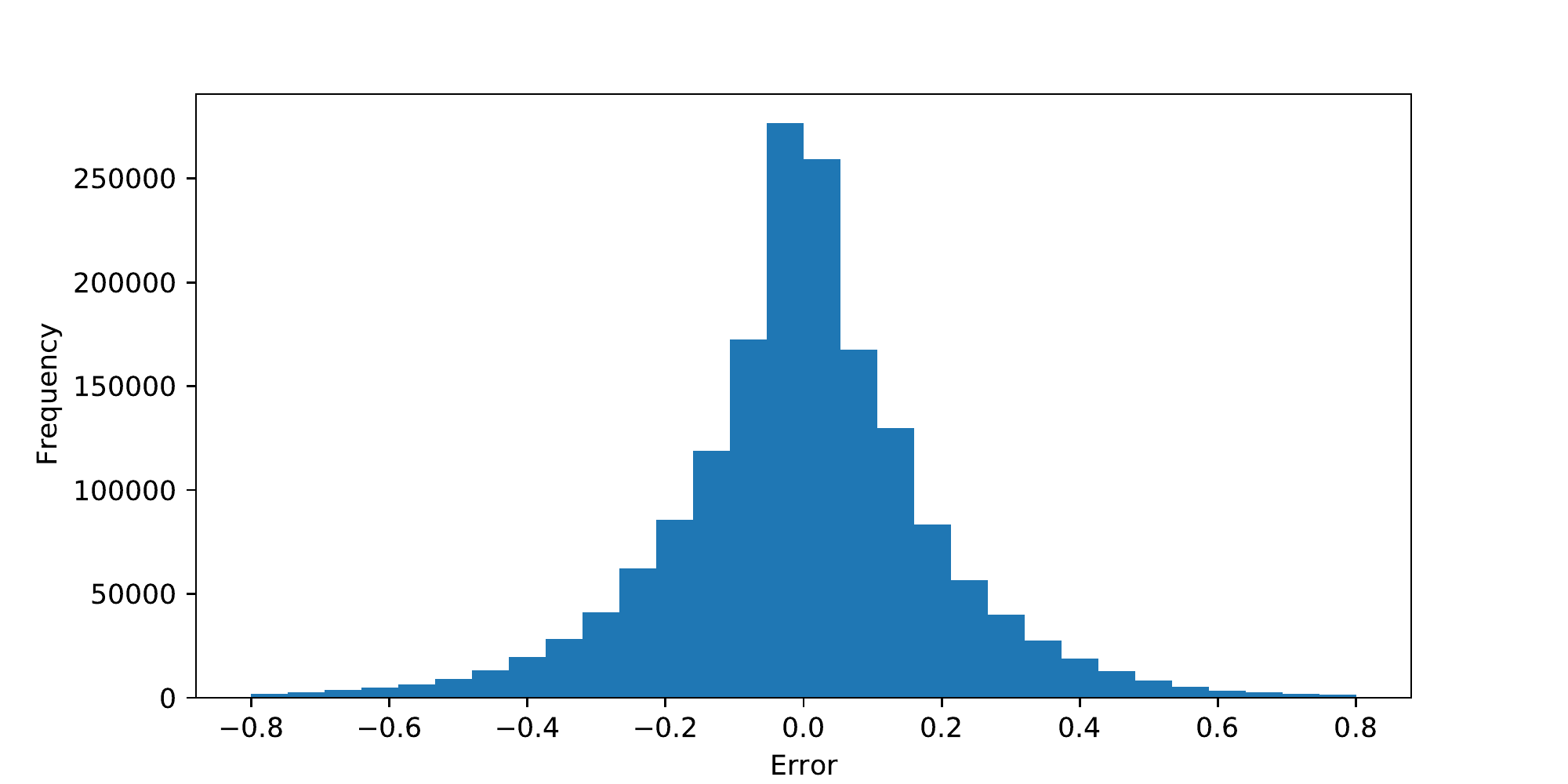}  
    \caption{\textbf{Histogram of SPIN axis-angle errors.} On 3DPW train set, in radians.}
    \label{fig:error_gaussian}
\end{figure}

\begin{figure*}
\centering
\includegraphics{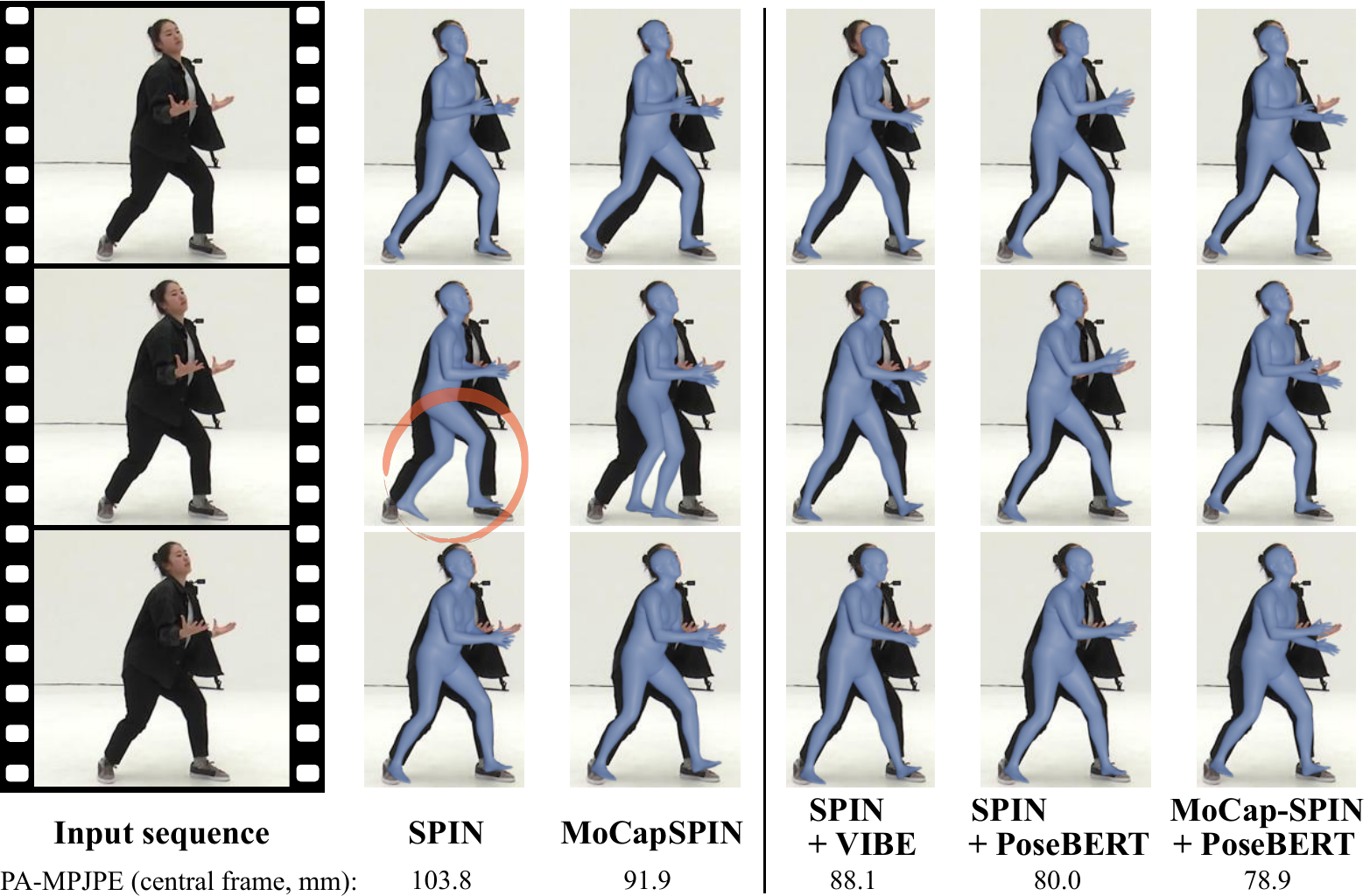}
\caption{\label{fig:extended_teaser}
\textbf{Qualitative comparison of image-based (left) and video-based (right) pose estimation methods} (extension of paper's Figure 1). Note the left/right legs flipping produced by SPIN, and how PoseBERT improves predictions by leveraging temporal information and MoCap-based motion priors.
}
\end{figure*}

We see this as a form of data augmentation for the axis-angle pose representation. We utilized histograms of axis-angle errors in radians shown in Figure \ref{fig:error_gaussian}
to estimate the standard deviation of the noise to be added (we used 0.10 to cover the error distribution) and simply sample a high-dimensional noise vector and add it to the ground truth input.

\setlength{\columnsep}{10pt}%

\subsection{Comparison on the use of 3DPW-train}

In the main paper we follow the standard evaluation protocol of 3DPW \cite{3dpw} which does not allow using 3DPW-train for training.
However we also provide in Table \ref{tab:3dpw_train} a comparison against methods using 3DPW-train for training.
We do not get any improvement by incorporating 3DPW into our training set but we still get results on part with other methods.

\begin{table}[h]
\vspace{-0.1cm}
\centering
\resizebox{\linewidth}{!}{
\begin{tabular}{lcccc}
\toprule
\multirow{2}{*}{Method} & Train set for & \multirow{2}{*}{Conf} & \multirow{2}{*}{3DPW} & MPI- \\
 & \textbf{the temporal module} & & &INF-3DHP\\
\midrule 
SPIN     & - & ICCV'19 & 59.2 & 67.5 \\
\bf{\sspin} & - & (ours) & 55.6 & 66.7 \\
\midrule
SPIN + VIBE & 2/3D & CVPR'20 & 56.5 & 63.4 \\
SPIN + MEVA & 2/3D+AMASS\textbf{+3DPW} & ACCV'20 & 54.7 & 65.4 \\
SPIN + TCMR & 2/3D & CVPR'21 & 55.8 & 62.8 \\
SPIN + TCMR & 2/3D\textbf{+3DPW} & CVPR'21 & 52.7 & 63.5 \\
\bf{SPIN + \posebert} & only synthetic (AMASS) & (ours) & 57.6 & 64.3 \\
\bf{\sspin + \posebert} & only synthetic (AMASS) & (ours) & 52.9 & 63.3 \\
\bf{\sspin + \posebert} & AMASS+\textbf{3DPW} & (ours) & 52.9 & 63.3 \\
\bottomrule
\end{tabular}
}
\caption{
\textbf{Comparison with state-of-the-art video models.}
"2D/3D" corresponds to the mix of MPI-INF-3DHP/PennAction/PoseTrack/H36M datasets.
}
\label{tab:3dpw_train}
\end{table}

\subsection{PoseBERT as a plug-and-play module}
Here, we further present results when running PoseBERT on top of the more recent ROMP method \cite{ROMP}.
Adding PoseBERT on top, we get a decrease of MPJPE (resp. PA-MPJPE) from 91.1 (resp. 56.5) to 90.2 (resp. 55.3) for 3DPW (using the cropping strategy of SPIN).
Finally, to test it beyond SMPL inputs, we appended PoseBERT on top of 2D/3D pose predictions from LCRNet++ \cite{lcrnet++} and see significant gains on 3DPW, a decrease of MPJPE (resp. PA-MPJPE) from 125.8 (resp. 68.8) to 108.2 (resp. 58.5).
We believe that these additional results further strengthen our argument that PoseBERT is a robust, plug-and-play 
module
for temporal modeling.

%% file: main.bbl
\begin{thebibliography}{10}\itemsep=-1pt

\bibitem{mpii}
Mykhaylo Andriluka, Leonid Pishchulin, Peter Gehler, and Bernt Schiele.
\newblock {2D} human pose estimation: New benchmark and state of the art
  analysis.
\newblock In {\em CVPR}, 2014.

\bibitem{videokinetics}
Anurag Arnab, Carl Doersch, and Andrew Zisserman.
\newblock Exploiting temporal context for {3D} human pose estimation in the
  wild.
\newblock In {\em CVPR}, 2019.

\bibitem{smplify}
Federica Bogo, Angjoo Kanazawa, Christoph Lassner, Peter~V. Gehler, Javier
  Romero, and Michael~J. Black.
\newblock Keep it {SMPL:} automatic estimation of {3D} human pose and shape
  from a single image.
\newblock In {\em ECCV}, 2016.

\bibitem{simclr}
Ting Chen, Simon Kornblith, Mohammad Norouzi, and Geoffrey Hinton.
\newblock A simple framework for contrastive learning of visual
  representations.
\newblock In {\em ICML}, 2020.

\bibitem{ChenWLSWTLCC16}
Wenzheng Chen, Huan Wang, Yangyan Li, Hao Su, Zhenhua Wang, Changhe Tu, Dani
  Lischinski, Daniel Cohen{-}Or, and Baoquan Chen.
\newblock Synthesizing training images for boosting human {3D} pose estimation.
\newblock In {\em {3DV}}, 2016.

\bibitem{tcmr}
Hongsuk Choi, Gyeongsik Moon, and Kyoung~Mu Lee.
\newblock Beyond static features for temporally consistent {3D} human pose and
  shape from a video.
\newblock {\em arXiv preprint arXiv:2011.08627}, 2020.

\bibitem{pose2mesh}
Hongsuk Choi, Gyeongsik Moon, and Kyoung~Mu Lee.
\newblock Pose2mesh: Graph convolutional network for {3D} human pose and mesh
  recovery from a 2d human pose.
\newblock In {\em ECCV}, 2020.

\bibitem{bert}
Jacob Devlin, Ming-Wei Chang, Kenton Lee, and Kristina Toutanova.
\newblock Bert: Pre-training of deep bidirectional transformers for language
  understanding.
\newblock {\em arXiv preprint arXiv:1810.04805}, 2018.

\bibitem{sim2real}
Carl Doersch and Andrew Zisserman.
\newblock Sim2real transfer learning for {3D} human pose estimation: motion to
  the rescue.
\newblock In {\em NeurIPS}, 2019.

\bibitem{frankle2020training}
Jonathan Frankle, David~J Schwab, and Ari~S Morcos.
\newblock Training batchnorm and only batchnorm: On the expressive power of
  random features in cnns.
\newblock In {\em ICLR}, 2021.

\bibitem{GaidonLP18}
Adrien Gaidon, Antonio~M. L{\'{o}}pez, and Florent Perronnin.
\newblock The reasonable effectiveness of synthetic visual data.
\newblock {\em IJCV}, 2018.

\bibitem{huang2017}
Yinghao Huang, Federica Bogo, Christoph Lassner, Angjoo Kanazawa, Peter~V.
  Gehler, Ijaz Akhter, and Michael~J. Black.
\newblock Towards accurate markerless human shape and pose estimation over
  time.
\newblock In {\em {3DV}}, 2017.

\bibitem{batchnorm}
Sergey Ioffe and Christian Szegedy.
\newblock Batch normalization: Accelerating deep network training by reducing
  internal covariate shift.
\newblock In {\em ICML}, 2015.

\bibitem{h36}
Catalin Ionescu, Dragos Papava, Vlad Olaru, and Cristian Sminchisescu.
\newblock {Human3.6M}: Large scale datasets and predictive methods for {3D}
  human sensing in natural environments.
\newblock {\em IEEE Trans. PAMI}, 2013.

\bibitem{Jiang_2021_CVPR}
Tao Jiang, Necati~Cihan Camgoz, and Richard Bowden.
\newblock Skeletor: Skeletal transformers for robust body-pose estimation.
\newblock In {\em Proceedings of the IEEE/CVF Conference on Computer Vision and
  Pattern Recognition (CVPR) Workshops}, June 2021.

\bibitem{lsp}
Sam Johnson and Mark Everingham.
\newblock Clustered pose and nonlinear appearance models for human pose
  estimation.
\newblock In {\em BMVC}, 2010.

\bibitem{hmr}
Angjoo Kanazawa, Michael~J Black, David~W Jacobs, and Jitendra Malik.
\newblock End-to-end recovery of human shape and pose.
\newblock In {\em CVPR}, 2018.

\bibitem{hmmr}
Angjoo Kanazawa, Jason~Y Zhang, Panna Felsen, and Jitendra Malik.
\newblock Learning {3D} human dynamics from video.
\newblock In {\em CVPR}, 2019.

\bibitem{adam}
Diederik~P Kingma and Jimmy Ba.
\newblock Adam: A method for stochastic optimization.
\newblock {\em arXiv preprint arXiv:1412.6980}, 2014.

\bibitem{vibe}
Muhammed Kocabas, Nikos Athanasiou, and Michael~J Black.
\newblock Vibe: Video inference for human body pose and shape estimation.
\newblock In {\em CVPR}, 2020.

\bibitem{spin}
Nikos Kolotouros, Georgios Pavlakos, Michael~J Black, and Kostas Daniilidis.
\newblock Learning to reconstruct {3D} human pose and shape via model-fitting
  in the loop.
\newblock In {\em {ICCV}}, 2019.

\bibitem{graphcmr}
Nikos Kolotouros, Georgios Pavlakos, and Kostas Daniilidis.
\newblock Convolutional mesh regression for single-image human shape
  reconstruction.
\newblock In {\em CVPR}, 2019.

\bibitem{kundu2020unsupervised}
Jogendra~Nath Kundu, Ambareesh Revanur, Govind~Vitthal Waghmare, Rahul~Mysore
  Venkatesh, and R~Venkatesh Babu.
\newblock Unsupervised cross-modal alignment for multi-person 3d pose
  estimation.
\newblock In {\em ECCV}, 2020.

\bibitem{unite}
Christoph Lassner, Javier Romero, Martin Kiefel, Federica Bogo, Michael~J
  Black, and Peter~V Gehler.
\newblock Unite the people: Closing the loop between {3D} and {2D} human
  representations.
\newblock In {\em CVPR}, 2017.

\bibitem{mc3dv}
Vincent Leroy, Philippe Weinzaepfel, Romain Br{\'e}gier, Hadrien Combaluzier,
  and Gr{\'e}gory Rogez.
\newblock {SMPLy} benchmarking {3D} human pose estimation in the wild.
\newblock In {\em 3DV}, 2020.

\bibitem{coco}
Tsung-Yi Lin, Michael Maire, Serge Belongie, James Hays, Pietro Perona, Deva
  Ramanan, Piotr Doll{\'a}r, and C~Lawrence Zitnick.
\newblock Microsoft coco: Common objects in context.
\newblock In {\em ECCV}, 2014.

\bibitem{smpl}
Matthew Loper, Naureen Mahmood, Javier Romero, Gerard Pons-Moll, and Michael~J.
  Black.
\newblock {SMPL}: a skinned multi-person linear model.
\newblock {\em ACM Transactions on Graphics}, 2015.

\bibitem{meva}
Zhengyi Luo, S~Alireza Golestaneh, and Kris~M Kitani.
\newblock 3d human motion estimation via motion compression and refinement.
\newblock In {\em ACCV}, 2020.

\bibitem{amass}
Naureen Mahmood, Nima Ghorbani, Nikolaus~F. Troje, Gerard Pons-Moll, and
  Michael Black.
\newblock {AMASS}: {Archive} of {Motion} {Capture} {As} {Surface} {Shapes}.
\newblock In {\em {ICCV}}, 2019.

\bibitem{mpiinf}
Dushyant Mehta, Helge Rhodin, Dan Casas, Pascal Fua, Oleksandr Sotnychenko,
  Weipeng Xu, and Christian Theobalt.
\newblock Monocular 3d human pose estimation in the wild using improved cnn
  supervision.
\newblock In {\em 3DV}, 2017.

\bibitem{mupots}
Dushyant Mehta, Oleksandr Sotnychenko, Franziska Mueller, Weipeng Xu, Srinath
  Sridhar, Gerard Pons-Moll, and Christian Theobalt.
\newblock Single-shot multi-person 3d pose estimation from monocular rgb.
\newblock In {\em 3DV}, 2018.

\bibitem{i2lmeshnet}
Gyeongsik Moon and Kyoung~Mu Lee.
\newblock I2l-meshnet: Image-to-lixel prediction network for accurate {3D}
  human pose and mesh estimation from a single rgb image.
\newblock {\em arXiv preprint arXiv:2008.03713}, 2020.

\bibitem{nbf}
Mohamed Omran, Christoph Lassner, Gerard Pons-Moll, Peter Gehler, and Bernt
  Schiele.
\newblock Neural body fitting: Unifying deep learning and model based human
  pose and shape estimation.
\newblock In {\em 3DV}, 2018.

\bibitem{pytorch}
Adam Paszke, Sam Gross, Francisco Massa, Adam Lerer, James Bradbury, Gregory
  Chanan, Trevor Killeen, Zeming Lin, Natalia Gimelshein, Luca Antiga, Alban
  Desmaison, Andreas Kopf, Edward Yang, Zachary DeVito, Martin Raison, Alykhan
  Tejani, Sasank Chilamkurthy, Benoit Steiner, Lu Fang, Junjie Bai, and Soumith
  Chintala.
\newblock Pytorch: An imperative style, high-performance deep learning library.
\newblock 2019.

\bibitem{texturepose}
Georgios Pavlakos, Nikos Kolotouros, and Kostas Daniilidis.
\newblock Texturepose: Supervising human mesh estimation with texture
  consistency.
\newblock In {\em ICCV}, 2019.

\bibitem{thmmr}
Georgios Pavlakos, Jitendra Malik, and Angjoo Kanazawa.
\newblock Human mesh recovery from multiple shots.
\newblock {\em arXiv preprint arXiv:2012.09843}, 2020.

\bibitem{RogezS16}
Gr{\'{e}}gory Rogez and Cordelia Schmid.
\newblock Mocap-guided data augmentation for {3D} pose estimation in the wild.
\newblock In {\em NIPS}, 2016.

\bibitem{lcrnet++}
Gregory Rogez, Philippe Weinzaepfel, and Cordelia Schmid.
\newblock {LCR-Net++: Multi-person {2D} and {3D} pose detection in natural
  images}.
\newblock {\em IEEE trans. PAMI}, 2020.

\bibitem{dct}
Yu Rong, Ziwei Liu, Cheng Li, Kaidi Cao, and Chen~Change Loy.
\newblock Delving deep into hybrid annotations for {3D} human recovery in the
  wild.
\newblock In {\em ICCV}, 2019.

\bibitem{schneider2020improving}
Steffen Schneider, Evgenia Rusak, Luisa Eck, Oliver Bringmann, Wieland Brendel,
  and Matthias Bethge.
\newblock Improving robustness against common corruptions by covariate shift
  adaptation.
\newblock {\em Advances in Neural Information Processing Systems}, 33, 2020.

\bibitem{ROMP}
Yu Sun, Qian Bao, Wu Liu, Yili Fu, Black Michael~J., and Tao Mei.
\newblock Monocular, one-stage, regression of multiple 3d people.
\newblock In {\em ICCV}, October 2021.

\bibitem{sun2019human}
Yu Sun, Yun Ye, Wu Liu, Wenpeng Gao, Yili Fu, and Tao Mei.
\newblock Human mesh recovery from monocular images via a skeleton-disentangled
  representation.
\newblock In {\em ICCV}, 2019.

\bibitem{aist}
Shuhei Tsuchida, Satoru Fukayama, Masahiro Hamasaki, and Masataka Goto.
\newblock Aist dance video database: Multi-genre, multi-dancer, and
  multi-camera database for dance information processing.
\newblock In {\em ISMIR}, 2019.

\bibitem{bodynet}
Gul Varol, Duygu Ceylan, Bryan Russell, Jimei Yang, Ersin Yumer, Ivan Laptev,
  and Cordelia Schmid.
\newblock Bodynet: Volumetric inference of 3d human body shapes.
\newblock In {\em ECCV}, 2018.

\bibitem{surreal}
G{\"u}l Varol, Javier Romero, Xavier Martin, Naureen Mahmood, Michael~J. Black,
  Ivan Laptev, and Cordelia Schmid.
\newblock Learning from synthetic humans.
\newblock In {\em CVPR}, 2017.

\bibitem{transformer}
Ashish Vaswani, Noam Shazeer, Niki Parmar, Jakob Uszkoreit, Llion Jones,
  Aidan~N Gomez, Lukasz Kaiser, and Illia Polosukhin.
\newblock In {\em NeurIPS}, 2017.

\bibitem{3dpw}
Timo von Marcard, Roberto Henschel, Michael~J. Black, Bodo Rosenhahn, and
  Gerard Pons-Moll.
\newblock Recovering {Accurate} {3D} {Human} {Pose} in the {Wild} {Using}
  {IMUs} and a {Moving} {Camera}.
\newblock In {\em {ECCV}}, 2018.

\bibitem{lsun}
Fisher Yu, Yinda Zhang, Shuran Song, Ari Seff, and Jianxiong Xiao.
\newblock {LSUN}: Construction of a large-scale image dataset using deep
  learning with humans in the loop.
\newblock {\em arXiv preprint arXiv:1506.03365}, 2015.

\bibitem{rotation}
Yi Zhou, Connelly Barnes, Jingwan Lu, Jimei Yang, and Hao Li.
\newblock On the continuity of rotation representations in neural networks.
\newblock In {\em CVPR}, 2019.

\bibitem{simpose}
Tyler Zhu, Per Karlsson, and Christoph Bregler.
\newblock Simpose: Effectively learning densepose and surface normals of people
  from simulated data.
\newblock In {\em ECCV}, 2020.

\end{thebibliography}
